%% file: colm2026_conference_preprint__1_.tex
\documentclass{article}

\usepackage[preprint]{colm2026_conference}

\usepackage{times}
\usepackage{latexsym}

\usepackage{inconsolata}

\input{math_commands.tex}

\usepackage[utf8]{inputenc} 
\usepackage[T1]{fontenc}    
\usepackage{url}            
\usepackage{booktabs}       
\usepackage{amsfonts}       
\usepackage{amssymb}        
\usepackage{nicefrac}       
\usepackage{microtype}      
\usepackage{xcolor}         
\usepackage{amsmath}
\usepackage{bbm} 
\usepackage{graphicx}
\usepackage{subcaption}
\usepackage{algorithm}
\usepackage{algorithmic}
\usepackage{wrapfig}
\usepackage{multirow}
\usepackage{pifont}
\usepackage{xcolor}
\usepackage{colortbl}   
\usepackage[most]{tcolorbox} 
\usepackage{tabularx}
\usepackage{xspace}
\usepackage{caption}  
\usepackage{makecell} 
\usepackage{colortbl}
\usepackage{enumitem}
\usepackage{cuted}
\usepackage[normalem]{ulem}

\newtheorem{theorem}{Theorem}[section]

\newtheorem{definition}[theorem]{Definition}

\usepackage{array}
\newcolumntype{L}[1]{>{\raggedright\let\newline\\\arraybackslash\hspace{0pt}}m{#1}}
\newcolumntype{C}[1]{>{\centering\let\newline  \\\arraybackslash\hspace{0pt}}m{#1}}
\newcolumntype{R}[1]{>{\raggedleft\let\newline \\\arraybackslash\hspace{0pt}}m{#1}}

\newcommand{\ours}{\textsc{PSPA-Bench}\xspace}

\newcommand{\cmark}{\textcolor{green}{\ding{51}}} 
\newcommand{\xmark}{\textcolor{red}{\ding{55}}}         

\title{\ours: A Personalized Benchmark for Smartphone GUI Agent}

\author{
 \textbf{Hongyi Nie\textsuperscript{1}},
 \textbf{Xunyuan Liu\textsuperscript{2}},
 \textbf{Yudong Bai\textsuperscript{3}},
 \textbf{Yaqing Wang\textsuperscript{4}},
 \textbf{Yang Liu\textsuperscript{1}},\\
 \textbf{Quanming Yao\textsuperscript{2}},
 \textbf{Zhen Wang\textsuperscript{1}}
\\
\\
\\
 \textsuperscript{1}Northwestern Polytechnical University,
 \textsuperscript{2}Tsinghua University,\\
 \textsuperscript{3}Peking University,
 \textsuperscript{4}Beijing Institute of Mathematical Sciences and Applications,
}

\begin{document}
\maketitle
\begin{abstract}
  Smartphone GUI agents operate directly on app interfaces, offering broad capability without deep system integration.  
  However, real-world use is highly personalized: users follow diverse workflows and preferences, demanding customized rather than generic assistance.
  Existing benchmarks overlook this personalization dimension due to limited user-specific data and the lack of fine-grained evaluation.
  To address this gap, we introduce \ours, a benchmark dedicated to evaluating personalization in smartphone GUI agents. Specifically,
  \ours utilizes the \emph{task decomposition graph, TDG} to represent the structure of personalized task at unit-instruction level. Using TDG, \ours generates task templates that instantiate personalized instructions without large-scale user logs, and supports fine-grained evaluation of GUI agents' persoalization abilities.
  \ours includes over 12,855 realistic personalized instructions spanning 10 scenarios and 22 commonly used mobile apps, along with four fine-grained evaluation metrics. We evaluate 11 state-of-the-art GUI agents and find that existing methods perform poorly under personalized settings, with even the best agent achieving limited success. Based on these results, we derive five insights to guide future research, positioning \ours as a foundation for advancing personalized GUI agents.
\end{abstract}

\section{Introduction}
\label{sec:introduction}
Smartphone GUI agents \citep{Zhang2025, Liu2025} have emerged as a promising paradigm for facilitating human-computer interaction in mobile environments. Different from conventional automation systems (e.g., Siri \footnote{\url{https://www.apple.com/siri/}}) that rely on system APIs, GUI agents \citep{Zhang2025a, Wang2024, Qin2025, Wen2024} operate directly on the graphical user interfaces of mobile applications. By emulating user actions such as tapping, swiping, and text input, these agents remain effective in ecosystems with limited or no backend integration.

\begin{figure*}[t]
   \centering
   \includegraphics[width=0.90\linewidth]{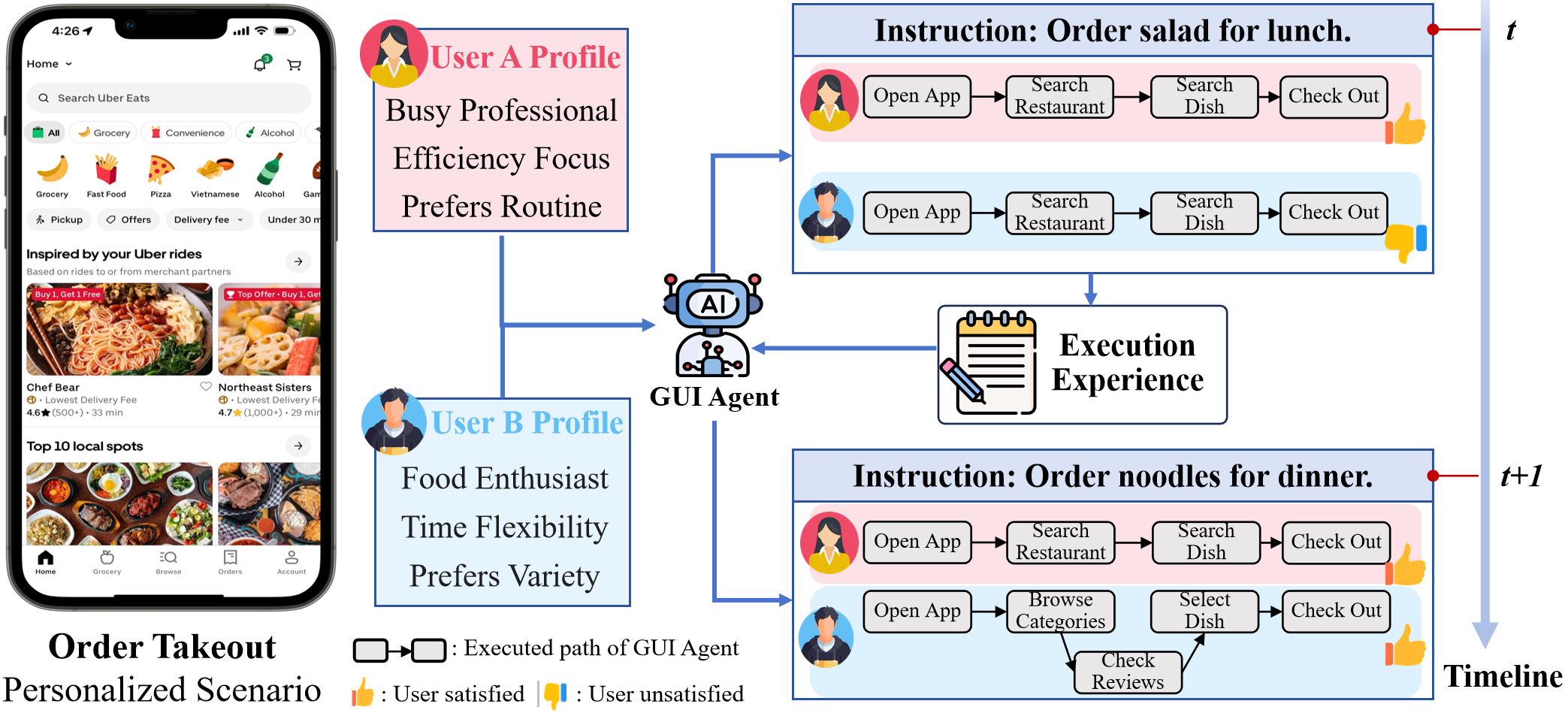}
   \caption{GUI agents are shifting from \textit{general-purpose, one-time} task execution to \textit{personalized, long-term} service. (i) Given the same instruction, an agent should adapt its execution path to user preferences; (ii) For the same user, the agent uses historical execution experience to provide long-term support; experience up to time $t$ informs a refined execution at $t+1$ for similar instructions.}
\label{fig:introduction}
\end{figure*}

However, unlike many other computing platforms, smartphone usage is inherently personalized \citep{Li2015, nguyen-etal-2025-gui}. As shown in Figure~\ref{fig:introduction}, users vary significantly in how they accomplish the same task through diverse action sequences \citep{Tian2020}. For instance, when ordering takeout, different users may prefer different search strategies and different food categories. Consequently, the demand for smartphone GUI agents is shifting away from \textit{generic, one-size-fits-all} solutions toward \textit{personalized, context-aware} assistants. Moreover, a GUI agent is not intended to act as a \textit{one-time} helper; rather, it is expected to provide continuous support over \textit{long-term use} \citep{Lee2024, Wang2025, Hu2025}. This requires the agent to accumulate execution experience and adapt to evolving individual preferences, delivering a ``more understanding of you'' interaction experience progressively.

Existing benchmarks evaluate GUI agents from different perspectives. For example, Mobile-Bench \citep{deng-etal-2024-mobile}, AndroidWorld \citep{Rawles2025} and Spa-Bench \citep{Chen2025}, focus on the general task-execution capabilities of GUI agents; TransBench \citep{lu-etal-2025-transbench} emphasize the transferability of GUI agents; AEIA-MN \citep{Chen2025a} and Mobile Safety Bench \citep{Lee2024a} focus on the safety of GUI agents. 
However, considering both the \textbf{sparsity of user data} and the \textbf{absence of suitable metrics for personalized evaluation}, the systematic assessment of GUI agents' personalization capabilities remains largely unexplored.

To address this gap, we introduce \ours, a benchmark for systematic study of personalization in smartphone GUI agents.
\ours uses the Task Decomposition Graph (TDG) to explicitly represent the structure of personalized GUI tasks. To compensate for sparse user data, \ours generates personalized instructions by building TDG-derived templates without relying on large-scale user logs. For fine-grained evaluation, \ours measures the ratio of completed unit instructions as the fine-grained metrics by through graph-trace alignment at the unit-instruction level.
\ours covers 10 daily-use scenarios (e.g., shopping, travel, dining) across 22 mobile apps, defines 100 user personas, and yields 12,855 personalized instructions. 
A detailed comparison with existing smartphone GUI benchmarks is provided in Appendix~\ref{sec:comparison_with_other_benchmarks}.
With the proposed \ours, we conduct comprehensive evaluations with 11 GUI agents, and the evaluation results from \ours demonstrate that there is still a long way for the development of personalized GUI agents. Furthermore, we highlight 5 key findings that provide directions for the development of personalized GUI agents: (i) effective \textbf{perception} underpins personalized instruction execution; (ii) strong \textbf{reasoning} is required to handle ambiguity and complex behaviors; (iii) personalization amplifies the trade-off between \textbf{accuracy and efficiency}; (iv) \textbf{persistent memory} is essential for long-term adaptation; and (v) \textbf{self-evolution} further enhances this adaptive capability.
To summarize, our contributions are as follows:

\begin{itemize}[leftmargin=*]
\item The problem of personalization in smartphone GUI agents is systematically formalized, highlighting the scarcity of personalized user data and the lack of evaluation metrics as core challenges.

\item We propose \ours, a benchmark with a data generation method for sparse user data and process-oriented evaluation metrics based on task decomposition graphs. \ours covers 10 representative daily scenarios, 22 third-party apps, 100 user personas, and 12,855 personalized instructions

\item 11 state-of-the-art GUI agents are benchmarked on \ours, and the results reveal that current methods perform poorly under personalized settings, and further distill 5 key directions for future research based on the evaluation results.
\end{itemize}

\section{Definition of Personalized GUI Agent Task}
\label{sec:definition_of_personalized_gui_agent_task}

Let $Q$ be the space of GUI task instructions, $E = (\gS, \gA)$ be a GUI environment that contains a set $\gS=\{s_i\}$ of observable interface states  and a set $\gA=\{a_i\}$ of executable GUI actions  for state transitions. The definition of Personalized GUI Agent task is given as follows:

\begin{definition}
   \textbf{Personalized GUI Agent Task.}
   For a user $u$, the profile $\rho_u$ induces a conditional distribution $\mathcal{D}^{\rho_u}_Q$ over the instruction space $Q$. At each time step $t$, a personalized instruction $q_t^u$ is sampled from $\mathcal{D}^{\rho_u}_Q$. The agent operates in the GUI environment $E = (\gS, \gA)$ under a user-specific policy $\pi_u$ with two objectives: (i) \textbf{Immediate objective:} generate an action sequence that drives the environment $E$ toward the target states satisfying the current instruction $q_t^u$; (ii) \textbf{Long-term objective:} accumulate execution experience $\mathcal{H}_{u}$ to update $\pi_u$, thereby improving expected performance on future personalized instruction sequences sampled from $\mathcal{D}^{\rho_u}_Q$.
\end{definition}

\section{Benchmarking Setup}

Despite rapid progress in GUI agents, the systematic assessment of their personalization capabilities remains largely unexplored. 
Current benchmarks face two main limitations. 
First, privacy regulations (e.g., GDPR~\citep{gdpr2016}) and user concerns restrict the collection of user-specific data, hindering the creation of realistic yet privacy-compliant datasets.
Second, evaluations rely on coarse metrics such as binary success signals \citep{Chen2025, Rawles2025, Yang2025} or task completion rates \citep{Xing2024}, which assume fixed action paths and fail to assess adaptability to diverse user preferences or learning over time.

\begin{figure*}[t]
	\centering
	\includegraphics[width=1\linewidth]{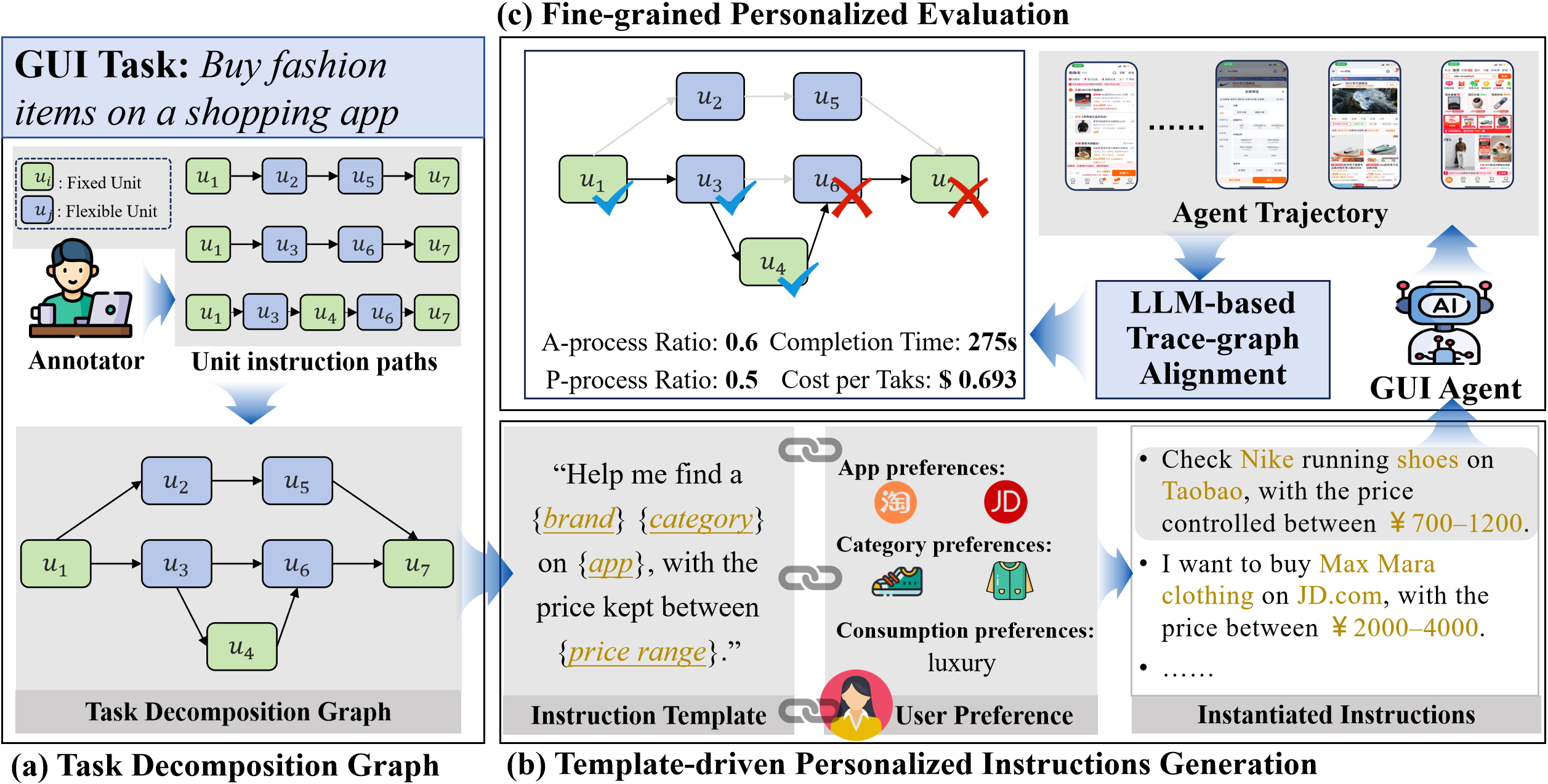}
	\caption{Benchmark framework of \ours.
    (a) \textbf{Task decomposition graph}: the task is decomposed into a directed acyclic graph of unit instructions, where fixed nodes denote universal steps and flexible nodes denote user-specific requirements.
		(b) \textbf{Template-driven personalized instruction generation}: 
    the TDG is used to construct task templates, which are then instantiated with user preferences to generate personalized instructions.
		(c) \textbf{Fine-grained personalized evaluation}: The trace-graph alignment is used to compute APR (A-Progress Ratio) and PPR (P-Progress Ratio) for fine-grained evaluation of both immediate and long-term performance.
	} 
	\label{fig:benchmark_setup}
\end{figure*}

To overcome these limitations, we propose \ours \footnote{The code and data of \ours are available at \url{https://anonymous.4open.science/status/PSPA-Bench}}, a benchmark specifically designed for personalized GUI agents (Figure~\ref{fig:benchmark_setup}) . 
At the core of \ours is the \emph{task decomposition graph} (TDG), which  explicitly captures the personalized structure of GUI agent tasks.
Based on the TDG, we explain how \ours generate personalized task instructions without relying on large-scale user logs, and evaluate agent executions with unit-instruction level fine-grained metrics. Finally, we present the comparison GUI agents employed in our benchmark.

\subsection{Task Decomposition Graph}

To address two limitations identified above, i.e., the sparsity of real user data and the lack of metrics tailored to personalization, we introduce the task decomposition graph (TDG). The TDG explicitly captures the personalized structure of GUI tasks. This structure (i) enables controlled generation of personalized task instances even when user traces are scarce, and (ii) supports fine-grained, preference-aware evaluation beyond coarse end-state signals.

Specifically, GUI tasks such as "Buy a piece of clothing on a shopping app" implicitly combine universal steps with user-specific requirements (e.g., preferred product category or price range).
To make these requirements explicit, we represent each task as a directed acyclic task decomposition graph $\mathcal{G} = (\mathcal{U}, \mathcal{E})$, where nodes are minimal GUI actions and edges denote execution dependencies. Any source-to-sink path defines a valid task completion. To model personalization, we distinguish fixed nodes $\mathcal{U}_f$ for user-invariant steps (e.g., launching the app) and flexible nodes $\mathcal{U}_p$ for preference-dependent actions (e.g., selecting preferred categories).

Therefore, compared with vague task descriptions, the TDG provides a more explicit and structured representation of the task with fine-grained unit instructions. This graph-based representation can serve as a unified foundation for generating personalized instructions and for conducting fine-grained,  evaluation of agent behavior. The constructed TDGs are provided in Appendix \ref{sec:task_graph_construction}. 

\subsection{Template-Driven Personalized Instruction Generation}

To address user data sparsity, we use the TDG as a generative blueprint to create diverse, realistic personalized instructions without large-scale user logs. The task structure ensures universal steps are preserved while systematically varying preference-dependent details.

\paragraph{Template construction.}  
The TDG explicitly separates fixed nodes $\mathcal{U}_f$ from flexible nodes $\mathcal{U}_p$. We use the fixed nodes and dependencies $\mathcal{E}$ as a procedural backbone, and convert flexible nodes into parameterized slots. This template preserves the shared task structure while clearly indicating where personalization occurs.

Specifically, each template $\mathsf{TP}_\mathcal{G}$ encodes the fixed procedural backbone of the task (derived from $\mathcal{U}_f$) together with a slot schema $\mathcal{L} = \{l_{u_p}\}_{u_p\in \mathcal{U}_p}$ mapping each flexible node $u_p$ to a slot $l_{u_p}$. Every slot $l\in \mathcal{L}$ is annotated with a semantic type $\tau(l)$ (e.g., \texttt{product\_category}, \texttt{price\_range}) and a domain of admissible values. The textual backbone preserves the execution ordering implied by $\mathcal{E}$, while slots mark precisely where personalization should be injected. We illustrate an example of the template for the shopping scenario as below:
\begin{tcolorbox}[colback=gray!15, colframe=black!30, title=\textbf{Template Example for Shopping Scenario}, fontupper=\small]
  \label{prompt: template_example_for_shopping_scenario}
  \small
  \texttt{Launch the shopping app, then search for a \underline{<product\_category>} within \underline{<price\_range>}, and add the top result to the cart.}
\end{tcolorbox} 
Here, ``<product\_category>'' and ``<price\_range>'' are slots derived from flexible nodes in the TDG, while the rest reflects the fixed structure.

\paragraph{Slot instantiation.}
Once a template is selected, we instantiate it by filling each parameterized slot with values sampled from the user’s preference distribution. For a user profile $\rho$, each slot $l$ is assigned a value $v_l \sim D_{\tau(l)}^{\rho}$, yielding a concrete instruction
$f(\mathsf{TP}_\mathcal{G}, {v_l}) \mapsto I$
that respects the dependencies in $\mathcal{G}$.
In this framework, the user profile $\rho$ directly controls the \emph{personalization strength}. 
A set of similar user profiles induces low-diversity slot values and thus weaker personalization, whereas a set of diverse profiles leads to more varied instructions, corresponding to stronger personalization signals. We provide the detailed instruction instantiation process in Appendix \ref{sec:instruction_instantiation}.

\paragraph{Task difficulty control.}  
To control task difficulty, we vary instruction templates along two dimensions:
(i) \textit{Complexity level} (low, medium, high), based on the number of unit instructions in the corresponding TDG.
(ii) \textit{Clarity level} (high, medium, low), which governs how explicitly the instruction communicates its intent.
We provide examples of templates in personalized scenarios in Appendix~\ref{sec:templates_of_personalized_gui_instruction}


\subsection{Fine-grained Personalized Evaluation}

We evaluate agents along two complementary objectives in fine-grained level (Section~\ref{sec:introduction}):  
(i) an \textit{immediate objective}: accurately and efficiently executing the current instruction, and  
(ii) a \textit{long-term objective}: adapting to user-specific preferences over long-term use.

Let the trace be $\mathcal{T} = (a_1,\ldots,a_L)$, where $a_i$ denotes the $i$-th action.
Since a TDG may contain multiple valid source-to-sink paths, we identify the path $P \subseteq \mathcal{G}$ that best matches the trace using a two-stage alignment procedure. First, we define an LLM-based alignment method $m:\mathcal{T}\rightarrow \mathcal{U}\cup\{\varnothing\}$ that assigns each action either to the unit instruction it fulfills or to $\varnothing$ if it does not correspond to any instruction. Second, we select the optimal path $P^*$ from all valid paths $\mathcal{P}$ in the TDG by maximizing the number of matched unit instructions:
\begin{equation}
P^* = \argmax_{P \in \mathcal{P}} \sum_{u \in P} \mathbbm{1}\left[\exists\, a_i \in \mathcal{T}: m(a_i) = u\right]
\label{eq:path_selection}
\end{equation}
where $\mathbbm{1}[\cdot]$ is the indicator function. 

\begin{wrapfigure}[13]{r}{0.35\textwidth}
  \centering
    \includegraphics[width=1\linewidth]{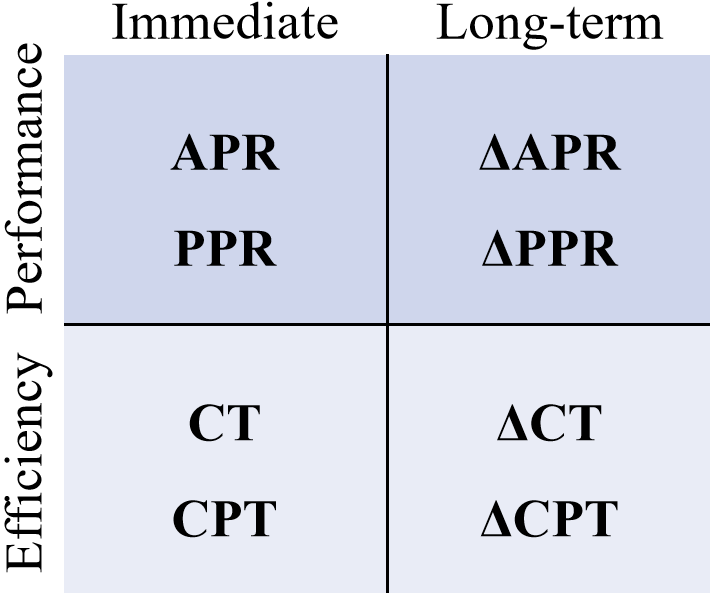}
    \caption{Evaluation metrics taxonomy of \ours.}
    \label{fig:metric_dimension}
\end{wrapfigure}
In case of ties (multiple paths achieving the same maximum match count), we apply the following tie-breaking rules in order: (i) prefer paths with more completed \textit{flexible} nodes $\mathcal{U}_p$, as they better reflect personalization; (ii) prefer shorter paths (fewer total nodes), favoring efficiency; (iii) if ties persist, select the path whose matched nodes appear earliest in the trace, reflecting the agent's primary execution intent.
Our evaluation begins by aligning the agent's execution trace with the corresponding TDG.
The resulting \emph{trace--graph} alignment indicates (i) which unit instructions were successfully completed and (ii) where failures occurred, with attribution to fixed nodes $\mathcal{U}_f$ (universal steps) and flexible nodes $\mathcal{U}_p$ (preference-sensitive steps).
Building on this, we assess these objectives along two metric dimensions (Figure~\ref{fig:metric_dimension}):  
\begin{itemize}[leftmargin=*]
  \item \textbf{Performance} measures execution quality at the unit-instruction level:  
  \textbf{A-process Ratio (APR)}: completion ratio over all unit instructions on the executed path.  
  \textbf{P-process Ratio (PPR)}: completion ratio restricted to flexible nodes $\mathcal{U}_p$, directly reflecting personalization.  
  \item  \textbf{Efficiency} quantifies the resources required for task execution:
  \textbf{Completion Time (CT)}: measured as the average elapsed time needed to finish a task.
  \textbf{Cost Per Task (CPT)}, which denotes the average monetary cost (USD) of completing a task, estimated from the number of model tokens consumed.
  \end{itemize}

Each metric is paired with an incremental variant `$\Delta$' that measures improvements over time, thereby capturing the agent's ability to adapt.  
For example, $\Delta$APR = APR$_{T_2} -$ APR$_{T_1}$ denotes performance gains between task $T_1$ and $T_2$, with analogous increments defined for PPR, CT, and CPT.  Implementation details of evaluation are provided in Appendix~\ref{sec:evaluation_details}.

\begin{table*}[t]
  \caption{Comparison of evaluated methods in \ours. The table presents baseline LLMs, general agent frameworks, and GUI agent frameworks, assessed across five key dimensions: Perception, Reasoning, Memory, Evolution, and Collaboration.}
  \label{tab:comparison_methods}
  \setlength\tabcolsep{1pt}
  \centering
  \footnotesize
  \vspace{-10pt}
  \begin{tabular}{@{}cc c C{1cm} C{1cm} C{1cm} C{1cm} C{1cm} c C{1cm} C{1cm} C{1cm} @{}}
     \toprule
     \multicolumn{2}{c}{\textbf{Method}} & \multicolumn{2}{c}{\textbf{Perception}} & \multicolumn{2}{c}{\textbf{Reasoning}} & \multicolumn{2}{c}{\textbf{Memory}} & \multicolumn{2}{c}{\textbf{Evolution}} & \multicolumn{2}{c}{\textbf{Collaboration}} \\
     \cmidrule(r){1-2} \cmidrule(r){3-4} \cmidrule(r){5-6} \cmidrule(r){7-8} \cmidrule(r){9-10} \cmidrule(r){11-12}
     \textbf{Type} & \textbf{Name} & Text & Vision & Planning & Reflection & Temporary & Persistent & Rules & Toolkits & Single & Multiple \\\midrule     
      \multirow{2}{*}{\begin{tabular}[c]{@{}c@{}}\\ LLM\end{tabular}}                & Base    &               & \ding{51}            &              &                & \ding{51}             &             &              &                & \ding{51}                &                \\
                                                                                           & +A11y         &\ding{51}               &              &              &                & \ding{51}             &             &              &                & \ding{51}                &                \\
                                                                                           & +A11y+SoM         & \ding{51}           & \ding{51}             &              &                &  \ding{51}            &             &              &                & \ding{51}                &                \\ 
                                                                                           & +A11y+Thinking         & \ding{51}           &             & \ding{51}             &                &  \ding{51}            &             &              &                & \ding{51}                &                \\ \midrule
\multirow{2}{*}{\begin{tabular}[c]{@{}c@{}}General \\ Agent\\ Framework\end{tabular}} & ReAct           & \ding{51}              & \ding{51}             & \ding{51}            &                & \ding{51}            &             &              &                & \ding{51}                &                \\
                                                                                           & Self-Refine & \ding{51}              & \ding{51}             &            & \ding{51}              & \ding{51}            &           &              &                &                  &              \\ 
                                                                                           & Reflexion    & \ding{51}              & \ding{51}               & \ding{51}            & \ding{51}              & \ding{51}            & \ding{51}           &              &                &                  & \ding{51}              \\ \midrule
                                                                                          
     \multirow{4}{*}{\begin{tabular}[c]{@{}c@{}}GUI\\ Agent\\ Framework\end{tabular}}     
     & AppAgent   & \ding{51}             & \ding{51}             & \ding{51}            &               & \ding{51}            &           &            &               & \ding{51}                   &             \\
    
     & M3A          & \ding{51}             & \ding{51}             & \ding{51}            &                & \ding{51}            &             &              &                & \ding{51}                &                \\
                                                                                           & Mobile-AgentV2  &               & \ding{51}             & \ding{51}            & \ding{51}              & \ding{51}            & \ding{51}           &              &                &                  & \ding{51}              \\
                                                                                           & Mobile-Agent E  &               & \ding{51}             & \ding{51}            & \ding{51}              & \ding{51}            & \ding{51}           & \ding{51}            & \ding{51}              &                  & \ding{51}              \\ \bottomrule
     \end{tabular}%
\end{table*}

\subsection{Comparison Methods}

As shown in Table~\ref{tab:comparison_methods}, \ours benchmarks a wide range of baselines and recent agent frameworks, summarized along five dimensions: \textbf{Perception}, \textbf{Reasoning}, \textbf{Memory}, \textbf{Evolution}, and \textbf{Collaboration}. \textbf{Perception} refers to how agents sense the GUI, either via accessibility (A11y) trees or visual inputs. \textbf{Reasoning} covers planning and reflection. \textbf{Memory} distinguishes temporary memory (within a task) from persistent memory (reused across tasks). \textbf{Evolution} considers whether behaviors or toolkits can be updated, and \textbf{Collaboration} characterizes the number and roles of agents.

For the \textbf{LLM category}, LLM-Base is a standard LLM (e.g., GPT-4o-mini). LLM+A11y augments it with A11y trees, while LLM+A11y+SoM adds set of marks (SoM) in screenshots for stronger visual grounding. LLM+A11y+Thinking further enhances reasoning with advanced models (e.g., GPT-O3-Mini).  
For \textbf{general agent frameworks}, we include ReAct \citep{Yao2023}, which interleaves reasoning and action; Self-Refine \citep{Madaan2023}, which iteratively improves via feedback; and Reflexion \citep{Shinn2023}, which uses reflection memory across tasks.  
For the \textbf{GUI agent frameworks}, \ours evaluates four representatives: AppAgent \citep{Zhang2025a}, which uses multimodal grounding through GUI parsing and screenshots; M3A \citep{Rawles2025}, a GUI-specialized ReAct; Mobile-Agent V2 \citep{Wang2024}, a multi-agent system with planning, decision, and reflection; and Mobile-Agent E \citep{Wang2025}, which extends V2 with hierarchical collaboration, persistent memory, and self-evolution.

\section{Empirical Results}


\begin{table}[t]
  \caption{Mean values of evaluation metrics (APR, PPR, CT, CPT) across five scenarios under high clarity conditions, evaluated at three levels of task complexity (low, middle, high). The results reflect each method's capability in achieving the immediate objective of personalized GUI tasks, i.e., accurately and efficiently executing the current instruction}
  \vspace{-10pt}
  \label{tab:immediate_objective}
  \resizebox{\textwidth}{!}{%
  \begin{tabular}{@{}l|cccc|cccc|cccc@{}}
  \toprule
  \multirow{3}{*}{Method} & \multicolumn{12}{c}{Setting: Clarity fixed at middle, varying complexity} \\ \cmidrule(l){2-13} 
                          & \multicolumn{4}{c|}{Complexity: low}      & \multicolumn{4}{c|}{Complexity: middle}   & \multicolumn{4}{c}{Complexity: high} \\ \cmidrule(l){2-13} 
                          & APR$\uparrow$ & PPR$\uparrow$ & CT$\downarrow$ & CPT$\downarrow$ 
                          & APR$\uparrow$ & PPR$\uparrow$ & CT$\downarrow$ & CPT$\downarrow$ 
                          & APR$\uparrow$ & PPR$\uparrow$ & CT$\downarrow$ & CPT$\downarrow$ \\ \midrule
  LLM-Base                & 0.002 & 0     & 236.7 & 0.612  & 0.001 & 0    & 256.6 & 0.501  & 0.002 & 0     & 293.4 & 0.816  \\
  +A11y                   & 0.625 & 0.586 & \cellcolor{red!10}{176.7} & \cellcolor{red!10}{0.056} & 0.431 & 0.449 & \cellcolor{red!10}{189.0} & \cellcolor{red!10}{0.038}  & 0.364 & 0.356 & \cellcolor{red!10}{229.3} & \cellcolor{red!10}{0.062}  \\
  +A11y+SoM               & 0.643 & 0.596 & 243.6 & 0.671 & 0.463 & 0.461 & 254.3 & 0.525 & 0.373 & 0.372 & 293.3 & 0.930  \\
  +A11y+Thinking          & \cellcolor{red!10}{0.674} & \cellcolor{red!10}{0.624} & 197.7 & 0.673 & \cellcolor{red!10}{0.466} & \cellcolor{red!10}{0.464} & 233.0 & 0.726  & \cellcolor{red!10}{0.382} & \cellcolor{red!10}{0.378} & 274.8 & 0.884        \\ \midrule
  ReAct                   & \cellcolor{blue!10}{0.683} & \cellcolor{blue!10}{0.618} & \cellcolor{blue!10}{217.0} & 0.752 & \cellcolor{blue!10}{0.503} & \cellcolor{blue!10}{0.468} & \cellcolor{blue!10}{243.0} & \cellcolor{blue!10}{0.616} & 0.414 & 0.403 & \cellcolor{blue!10}{316.8} & \cellcolor{blue!10}{0.761}   \\
  Self-Refine             & 0.596 & 0.542 & 249.1 & 0.758 & 0.438 & 0.441 & 731.4 & 0.796 & 0.308 & 0.339 & 845.3 & 0.931  \\
  Reflexion               & 0.665 & 0.611 & 250.7 & \cellcolor{blue!10}{0.657} & 0.465 & 0.466 & 332.5 & 0.694 & \cellcolor{blue!10}{0.438} & \cellcolor{blue!10}{0.406} & 464.5 & 0.776  \\ \midrule
  AppAgent                & 0.654 & 0.592 & \cellcolor{green!10}{218.7} & \cellcolor{green!10}{0.329} & 0.469 & 0.495 & \cellcolor{green!10}{227.8} & \cellcolor{green!10}{0.337} & 0.373 & 0.402 & \cellcolor{green!10}{292.6} & \cellcolor{green!10}{0.502}  \\
  M3A                     & 0.685 & 0.623 & 317.6 & 0.951 & 0.521 & 0.485 & 340.7 & 1.010 & 0.460 & 0.403 & 450.6 & 1.507  \\
  Mobile-Agent V2         & 0.674 & 0.618 & 517.6 & 1.017 & 0.532 & 0.533 & 578.9 & 1.364 & 0.435 & 0.415 & 659.1 & 1.414  \\
  Mobile-Agent E          & \cellcolor{green!10}{0.695} & \cellcolor{green!10}{0.621} & 532.7 & 1.084 & \cellcolor{green!10}{0.546} & \cellcolor{green!10}{0.539} & 587.5 & 1.401 & \cellcolor{green!10}{0.468} & \cellcolor{green!10}{0.435} & 716.7 & 1.692  \\ \bottomrule
  \end{tabular}
  }
\end{table}

\subsection{Experiments on Immediate Objective}

We evaluate each method's ability to accomplish the immediate objective: executing a single personalized instruction effectively and efficiently. As shown in Table~\ref{tab:immediate_objective}, the experiment is conducted under high clarity conditions across three levels of task complexity (low, medium, high), and the evaluation is based on unit-level metrics (APR, PPR) and execution efficiency (CT, CPT).
The results reveal clear performance gaps between different model classes, shedding light on the capabilities required to handle personalized tasks. We summarize the key insights below.

(1) \textbf{Perception is the foundation for personalized instruction execution.}  
The drastic improvement from LLM-Base (APR = 0.002, PPR = 0) to +A11y (APR = 0.625, PPR = 0.586 at low complexity) highlights that access to GUI context is indispensable. Personalized tasks often rely on dynamic, user-specific content, such as preferred items, recent activity, or customized filters, that cannot be interpreted through language alone. Perception provides the necessary grounding, enabling the agent to map abstract instructions to the actual UI elements visible on the screen. Without this capability, the agent is effectively blind to the task's personalized components.

(2) \textbf{Reasoning is essential for resolving ambiguity and executing complex personalized behaviors.}  
Perception allows the agent to observe, but reasoning allows it to decide. Many personalized instructions contain implicit preferences, conditional steps, or multi-hop objectives (e.g., “find a deal that matches your usual budget and dietary preferences”). Reasoning-enhanced models such as +A11y+Thinking and ReAct show significant gains over perception-only agents as complexity increases. For instance, ReAct achieves APR = 0.503 and PPR = 0.468 at medium complexity, outperforming simpler baselines. These models can handle flexible execution paths, maintain coherence across steps, and adapt to user-specific constraints — all of which are critical in the context of personalization.

(3) \textbf{Personalization introduces a stronger trade-off between execution accuracy and efficiency.}  
While Mobile-Agent E achieves the highest APR (0.695) and PPR (0.621), it also exhibits the slowest execution (CT = 716.7 at high complexity), reflecting a strategy that prioritizes accuracy through exhaustive interaction. This behavior may stem from the need to explore interface options and disambiguate user preferences on the fly. In contrast, AppAgent adopts a more efficient but less exhaustive approach (CT = 292.6 at high complexity), with slightly lower accuracy. In personalized settings, where goals are less explicit and vary across users, striking the right balance between precision and efficiency becomes more challenging than in generic task settings.

\begin{table}[t]
  \caption{Mean values of evaluation metrics ($\Delta$APR, $\Delta$PPR, $\Delta$CT, $\Delta$CPT) across five scenarios, reported under two settings: (i) high clarity with varying task complexity (low, middle, high), and (ii) middle complexity with varying task clarity (low, middle, high). Before evaluation, each method performs personalized tasks with middle complexity and middle clarity to accumulate initial experience. The results reflect each method's capability in achieving the long-term objective of personalized GUI tasks, i.e., adapting to user preferences over long-term use. (Note: For presentation convenience, $\Delta$APR, $\Delta$PPR and $\Delta$CPT values are $\times 100$.)} 
  \label{tab:long_term_objective}
  \resizebox{\textwidth}{!}{%
  \footnotesize
  \begin{tabular}{@{}l|cccc|cccc|cccc@{}}
  \toprule
  \multirow{3}{*}{Method} 
      & \multicolumn{12}{c}{Setting (i): Clarity fixed at high, varying complexity} \\ \cmidrule(l){2-13} 
      & \multicolumn{4}{c|}{Complexity: low}      
      & \multicolumn{4}{c|}{Complexity: middle}   
      & \multicolumn{4}{c}{Complexity: high} \\ \cmidrule(l){2-13} 
      & $\Delta$APR$\uparrow$ & $\Delta$PPR$\uparrow$ & $\Delta$CT$\downarrow$ & $\Delta$CPT$\downarrow$ 
      & $\Delta$APR$\uparrow$ & $\Delta$PPR$\uparrow$ & $\Delta$CT$\downarrow$ & $\Delta$CPT$\downarrow$ 
      & $\Delta$APR$\uparrow$ & $\Delta$PPR$\uparrow$ & $\Delta$CT$\downarrow$ & $\Delta$CPT$\downarrow$     \\ \midrule
  LLM-Base                & 0      & 0       & 9.28 & 0.20    & 0      & 0       & -17.68 & 0.73 & 0      & 0      & -19.18 & 0.94 \\
  +A11y                   & 0.66 & -0.58 & -12.73 & -0.63 & -0.39 & 0.05 & -2.72 & 0.42 & 0.22 & -0.72 & -8.31 & -0.27 \\
  +A11y+SoM               & -0.09 & 0.57 & 12.01 & -0.03 & 0.18 & -0.91 & 4.30 & 0.66 & -0.87 & 0.90 & -18.63 & 0.62 \\
  +A11y+Thinking          & -0.39 & -0.80 & 7.37 & -0.12 & -0.76 & -0.01 & -18.62 & 0.82 & -0.48 & 0.33 & -7.53 & 0.04 \\ \midrule
  ReAct                   & 0.09 & -0.63 & -18.78 & 0.55 & 0.88 & 0.79 & -3.92 & 0.84 & -0.82 & -0.61 & -18.19 & 0.35 \\
  Self-Refine             & -0.22 & -0.46 & 13.15 & -0.29 & -0.44 & 0.09 & -14.36 & 0.60 & -0.85 & 0.97 & 10.89 & -0.60 \\
  \rowcolor{red!10} 
  Reflexion               & 4.53 & 3.71 & -38.35 & -2.46 & 3.54 & 3.29 & -40.62 & -2.77 & 2.90 & 1.25 & -26.76 & -1.87 \\ \midrule
  AppAgent                & -0.38 & -0.35 & -9.18 & 0.28 & 0.77 & -0.06 & -15.22 & 0.43 & 0.52 & 0.12 & 10.84 & -0.10 \\
  M3A                     & 0.05 & -0.14 & 18.98 & -0.78 & -0.94 & 0.27 & -7.43 & 0.02 & 0.82 & -0.50 & 3.58 & 0.51 \\
  \rowcolor{red!10} 
  Mobile-Agent V2         & 4.84 & 3.26 & -36.48 & -3.70 & 4.54 & 4.02 & -41.45 & -2.43 & 3.41 & 1.04 & -26.02 & -3.58 \\
  \rowcolor{red!10} 
  Mobile-Agent E          & 6.96 & 4.08 & -62.08 & -4.06 & 6.72 & 3.68 & -57.02 & -3.41 & 2.13 & 2.44 & -35.51 & -1.94 \\ \midrule
  \multicolumn{13}{c}{} \\[-0.8em] \midrule
  \multirow{3}{*}{Method} 
      & \multicolumn{12}{c}{Setting (ii): Complexity fixed at middle, varying clarity} \\ \cmidrule(l){2-13} 
      & \multicolumn{4}{c|}{Clarity: low}      
      & \multicolumn{4}{c|}{Clarity: middle}   
      & \multicolumn{4}{c}{Clarity: high} \\ \cmidrule(l){2-13} 
      & $\Delta$APR$\uparrow$ & $\Delta$PPR$\uparrow$ & $\Delta$CT$\downarrow$ & $\Delta$CPT$\downarrow$ 
      & $\Delta$APR$\uparrow$ & $\Delta$PPR$\uparrow$ & $\Delta$CT$\downarrow$ & $\Delta$CPT$\downarrow$ 
      & $\Delta$APR$\uparrow$ & $\Delta$PPR$\uparrow$ & $\Delta$CT$\downarrow$ & $\Delta$CPT$\downarrow$     \\ \midrule
  LLM-Base                & 0       & 0      & 17.72 & -0.35 & 0     & 0       & -5.45 & 0.94 & 0       & 0       & -0.11 & 0.40 \\
  +A11y                   & -0.43 & -0.93 & 4.38 & -0.01 & -0.90 & -0.44 & 16.33 & -0.52 & -0.71 & -0.02 & 19.43 & -0.52 \\
  +A11y+SoM               & 0.34 & 0.52 &  -10.49 & 0.46 & -0.26 & 0.26 & 5.34 & -0.07 & -0.82 & 0.67 & -7.17 & 0.63 \\
  +A11y+Thinking          & -0.92 & 0.18 & -7.10 &  0.97 & 0.02 & -0.55 & -5.81 & 0.65 & 0.38 & -0.23 & 17.47 & -0.72 \\ \midrule
  ReAct                   & -0.32 & -0.77 & 16.99 & -0.75 & -0.48 & 0.32 & -12.69 & 0.11 & 0.06 & -0.52 & -16.28 & 0.79 \\
  Self-Refine             & 0.80 & 0.27 & 6.44 & -0.30 & 0.45 & 0.79 & -15.48 & 0.56 & 0.28 & -0.83 & -13.53 & 0.80 \\
  \rowcolor{red!10} 
  Reflexion               & 4.23 & 4.58 & -37.50 & -2.75 & 3.70 & 2.75 & -39.54 & -2.78 & 2.11 & 1.71 & -31.34 & -1.33 \\ \midrule
  AppAgent                & -0.35 & 0.49 & 5.99 & -0.70 & 0.32 & 0.14 & -16.25 & 0.26 & -0.47 & -0.51 & 18.92 & -0.21 \\
  M3A                     & 0.78 & 0.26 & 11.79 & -0.65 & 0.15 & -0.01 & -12.19 & 0.44 & -0.44 & -0.95 & -5.82 & 0.65 \\
  \rowcolor{red!10} 
  Mobile-Agent V2         & 4.27 & 3.15 & -36.17 & -2.95 & 3.72 & 3.62 & -43.49 & -2.1 & 2.26 & 1.61 & -25.23 & -1.31 \\
  \rowcolor{red!10} 
  Mobile-Agent E          & 6.54 & 7.15 & -63.81 & -3.08 & 5.13 & 4.94 & -53.44 & -2.28 & 5.55 & 4.08 & -40.59 & -2.75  \\ \bottomrule
  \end{tabular}%
  }
\end{table}

\subsection{Experiments on long-term Objective}

We then evaluate each method's ability to achieve the long-term objective: adapting to user preferences through continued use. To simulate long-term interaction, agents are first exposed to tasks with medium complexity and medium clarity to accumulate execution experience. Their subsequent performance is compared against a baseline without such accumulation. We report the relative changes in $\Delta$APR, $\Delta$PPR, $\Delta$CT, and $\Delta$CPT, which are computed as the difference between task performance after experience accumulation and task performance without it. Evaluation is conducted under two conditions: (i) increasing task complexity with clarity fixed, and (ii) increasing ambiguity with complexity fixed. Table~\ref{tab:long_term_objective} summarizes the results, highlighting the critical role of memory and experience processing in supporting long-term personalization.  

Formally, let $M$ denote an evaluation metric (e.g., APR, PPR, CT, CPT). For a user $u$, we denote $M^{\text{with}}_u$ as the metric value after experience accumulation and $M^{\text{w/o}}_u$ as the metric value without accumulation. The relative improvement is defined as:
$\Delta M = \frac{1}{N} \sum_{u=1}^N (M^{\text{with}}_u - M^{\text{w/o}}_u )$,
where $N$ is the number of evaluation tasks. For APR and PPR, a positive $\Delta M$ indicates better task completion, while for CT and CPT, a negative $\Delta M$ indicates improved efficiency.
Table~\ref{tab:long_term_objective} reports the results, and we further highlight two key findings based on the evaluation results:

\textbf{(1) Persistent memory is the foundation for long-term adaptation.}  
Models without persistent memory, such as ReAct and M3A, show little to no improvement, and in many cases even regress (e.g., negative $\Delta$APR and $\Delta$PPR across most settings). In contrast, methods equipped with persistent memory, e.g., Reflexion and Mobile-Agent V2, achieve consistent gains. These results show that retaining and reusing past experience is indispensable for adapting to personalized, evolving user behaviors.

\textbf{(2) Self-evolution enhances long-term adaptation.}  
Persistent memory provides the foundation, but Mobile-Agent E goes further by enabling self-evolution: it can update execution rules and create shortcuts from accumulated experience in its persistent memory. This capability yields the strongest gains overall (e.g., $\Delta$APR = +6.72, $\Delta$PPR = +3.68 at medium complexity). Compared to Mobile-Agent V2, which mainly reuses stored memory, Mobile-Agent E actively improves its strategy over time, leading to better generalization to new but related personalized tasks.

\subsection{Ranking of Methods on Personalized GUI Benchmark}

\begin{figure}[t]
\vspace{-0.5cm}
\centering
\includegraphics[width=0.85\linewidth]{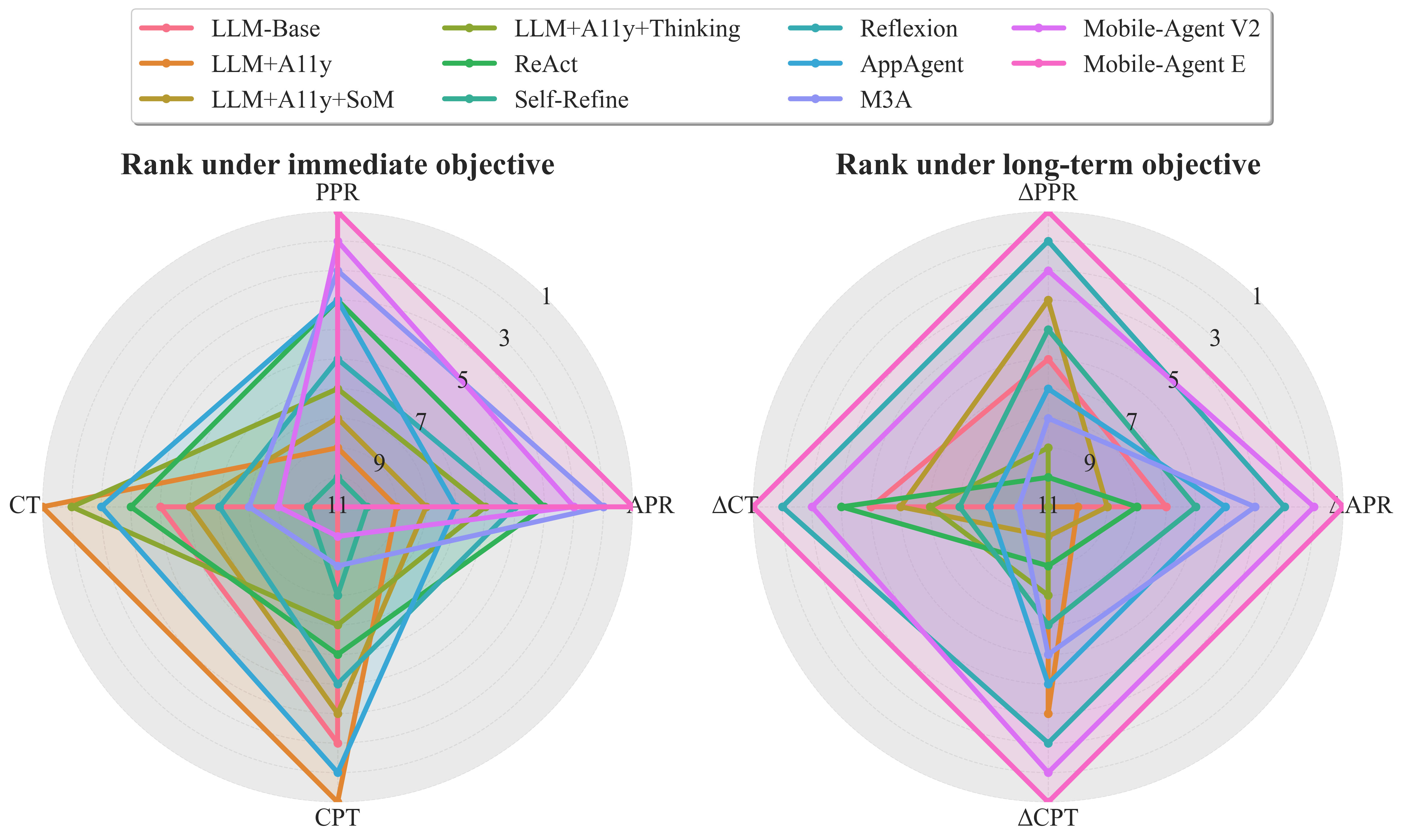}
\caption{Radar plots showing the comparative ranks of different methods under two objectives. 
\textbf{Left:} Immediate objective, evaluated with APR, PPR, CT, and CPT. 
\textbf{Right:} Long-term objective, evaluated with $\Delta$APR, $\Delta$PPR, $\Delta$CT, and $\Delta$CPT. 
Each axis indicates the rank of a method on the corresponding metric, with lower values representing better performance.}
\label{fig:rank_rank}
\vspace{-0.5cm}
\end{figure}

To further analyze the performance of different methods on the proposed \ours, 
we conduct a comprehensive ranking of methods based on the proposed \ours. 
We rank the methods based on the both the immediate objective and the long-term objective.  
As illustrated in Figure~\ref{fig:rank_rank}, for the immediate objective (left), 
a clear trade-off emerges between execution performance (APR, PPR) and efficiency (CT, CPT). 
Mobile-Agent E dominates in task completion accuracy but suffers from high execution cost, while AppAgent and LLM+A11y achieve faster execution with lower accuracy. 
This pattern confirms that the GUI agent need to balance the precision and efficiency for task execution.
For the long-term objective (right), methods with persistent memory, e.g., Reflexion, Mobile-Agent V2, and especially Mobile-Agent E, achieve consistently higher improvements across $\Delta$APR and $\Delta$PPR. This demonstrates that retaining and reusing past experiences is indispensable for adapting to evolving user preferences, with more advanced memory mechanisms further amplifying gains.

\section{Related Works}

\subsection{Smartphone GUI Agent}

Smartphone GUI agents \citep{Zhang2025,Liu2025} automate tasks by interpreting visual and textual elements on mobile screens. Prior work \citep{nguyen-etal-2025-gui} has explored several design dimensions. For perception, current agents rely on structured UI data such as A11y trees \citep{Rawles2025,Wen2024}, or use vision-based methods for richer context \citep{Zhang2025a,Gou2025}. For reasoning and planning, frameworks like ReAct \citep{Yao2023} and Reflexion \citep{Shinn2023} have been integrated into M3A \citep{Rawles2025} and Mobile-Agent V2 \citep{Wang2024} to handle complex tasks, while GUI-Explorer \citep{Xie2025} leverages prior knowledge of app operations. Recent studies also highlight learning from execution experience: Mobile-Agent E \citep{Wang2025} and AppAgentX \citep{Jiang2025} summarize repetitive low-level actions into high-level shortcuts to improve efficiency. Despite recent progress, most agents still target general automation, leaving personalization underexplored. Recently, FingerTip \citep{Chen2025} introduces a benchmark for evaluating personalization and shows that existing agents perform poorly on such tasks. However, its coarse-grained metrics may lose important details when assessing adaptation to diverse preferences or long-term use.

\subsection{Personalized LLM Agent}

Personalized LLM agents \citep{Dong2023, Tseng2024} aim to tailor assistance to individual preferences. Approaches fall into two paradigms: fine-tuning and training-free methods. Fine-tuning approaches, such as Personalized Soups \citep{jang2024personalized} and \citep{Liu2025a}, adapt model parameters using explicit user feedback. While effective in capturing preferences across multiple users, these methods require large amounts of data and significant computational resources. Training-free methods instead keep the base model fixed, incorporating personalization via prompt engineering \citep{Zhang2025b,Gao2024}, retrieval-augmented generation (RAG) \citep{Salemi2024,Salemi2024a}, or external memory \citep{Wang2024a,Zhong2024}. For example, CIPHER \citep{Gao2024} leverages user edits to infer preferences, while EMG-RAG \citep{Wang2024a} organizes user-specific information with editable memory graphs. While effective in static tasks such as travel planning, applying these strategies to dynamic mobile GUI environments remains an open challenge.

\vspace{-0.1cm}
\section{Conclusion}
We presented \ours, a benchmark for evaluating personalization in smartphone GUI agents. 
\ours introduces the TDG to explicitly separate universal and preference-sensitive steps, enabling both realistic personalized instruction generation and fine-grained evaluation. Covering 10 scenarios, 22 apps, and 100 user personas, \ours provides a large-scale, privacy-compliant testbed.  
Experiments with 11 representative models show that existing agents struggle on personalized tasks, revealing substantial room for improvement. 
At the same time, our analysis highlights that personalization requires agents to go beyond generic automation, emphasizing grounding in GUI context, structured reasoning, and mechanisms for accumulating and evolving user-specific experience. 
By providing a principled framework and comprehensive evaluation, \ours lays the foundation for advancing research on personalized GUI agents and offers a path toward more adaptive and user-centered mobile AI systems.

\clearpage

\section*{Limitations}

While \ours provides a broad benchmark for evaluating personalization in smartphone GUI agents, several limitations should be noted.
First, although the benchmark covers 22 apps across 10 representative scenarios, this is still a limited subset of the vast mobile application ecosystem, and the findings may not fully generalize to other domains such as healthcare, finance, or specialized professional tools.
Second, the user personas and preferences in \ours are synthetically generated based on predefined templates rather than collected from real users, which may not fully capture the complexity, inconsistency, and evolving nature of authentic user behaviors.
Third, the current benchmark is primarily designed for Android devices, limiting its direct applicability to other platforms such as iOS, where UI conventions and interaction patterns differ.
Finally, the task decomposition graphs are manually constructed by domain experts, which, while ensuring quality, may not cover all possible task variations or edge cases that arise in real-world usage.
These limitations also point to promising directions for future work, including extending the benchmark to broader application domains and platforms, incorporating real user data while preserving privacy, and developing automated methods for TDG construction.

\bibliographystyle{colm2026_conference}
\bibliography{PSPA}

\clearpage
\appendix
\section*{Appendix}

\section{Comparison with Other Benchmarks}
\label{sec:comparison_with_other_benchmarks}

\begin{table*}[htbp]
\centering
\caption{Comparison of General and Personalized GUI Benchmarks}
\label{tab:comparison_with_other_benchmarks}
\setlength\tabcolsep{5pt}
\resizebox{\textwidth}{!}{%
\begin{tabular}{cccccccccc}
\toprule
Type                                                                                     & \begin{tabular}[c]{@{}c@{}}Dataset \&\\ Benchmark\end{tabular} & \begin{tabular}[c]{@{}c@{}}Environment with\\ physical phones\end{tabular} & \begin{tabular}[c]{@{}c@{}}Third-part\\ apps\end{tabular} & \begin{tabular}[c]{@{}c@{}}User\\ profiles\end{tabular} & \begin{tabular}[c]{@{}c@{}}User\\ preferences\end{tabular} & \begin{tabular}[c]{@{}c@{}}Task\\ instructions\end{tabular} & \begin{tabular}[c]{@{}c@{}}Difficulty\\ dimensions\end{tabular}               & \begin{tabular}[c]{@{}c@{}}Fine-grained\\ progress evaluation\end{tabular} & \begin{tabular}[c]{@{}c@{}}Incremental\\ evaluation\end{tabular} \\ \midrule

\multirow{5}{*}{\begin{tabular}[c]{@{}c@{}}General\\ GUI\\ benchmark\end{tabular}} 
& AndroidArena     & \xmark & \xmark & \xmark & \xmark & 221    & \xmark & \cmark & \xmark \\
& LlamaTouch       & \xmark & \cmark & \xmark & \xmark & 495    & \begin{tabular}[c]{@{}c@{}}Step length\end{tabular} & \xmark & \xmark \\
& MobileAgentBench & \xmark & \xmark & \xmark & \xmark & 100    & \begin{tabular}[c]{@{}c@{}}Step length\end{tabular} & \xmark & \xmark \\
& AndroidWorld     & \xmark & \xmark & \xmark & \xmark & $\infty$ & \begin{tabular}[c]{@{}c@{}}Step length\end{tabular} & \xmark & \xmark \\
& SPA-bench        & \cmark & \cmark & \xmark & \xmark & 340    & \begin{tabular}[c]{@{}c@{}}Step length\end{tabular} & \xmark & \xmark \\
\midrule

\multirow{3}{*}{\begin{tabular}[c]{@{}c@{}}Personalized\\ GUI \\ benchmark\end{tabular}} 
& LearnGUI         & \xmark & \xmark & \xmark & \xmark & 2353   & \xmark & \xmark & \xmark \\
& FingerTip        & \cmark & \cmark & 91 & Fixed    & 21437  & \begin{tabular}[c]{@{}c@{}}Step length\end{tabular} & \xmark & \xmark \\
& \ours       & \cmark & \cmark & 100 & Dynamic  & 12855  & \begin{tabular}[c]{@{}c@{}}Step length \&\\ instruction clarity\end{tabular} & \cmark & \cmark \\
\bottomrule
\end{tabular}%
}
\end{table*}

Table~\ref{tab:comparison_with_other_benchmarks} summarizes the comparison between \ours and existing benchmarks for smartphone GUI agents. General-purpose GUI benchmarks such as AndroidArena, MobileAgentBench, and AndroidWorld are primarily designed to evaluate agents' basic task-execution capabilities. While they offer large-scale task datasets, they lack support for personalization—there are no user profiles, preferences, or long-term evaluation protocols. SPA-Bench improves realism by leveraging physical devices and third-party apps, but it still does not consider personalization.

Personalized GUI benchmarks, such as LearnGUI and FingerTip, take steps toward user-aware evaluation. However, LearnGUI is limited to simulated environments without physical devices or third-party apps, reducing ecological validity. FingerTip collects real interactions on physical devices with user profiles, but the user preferences are fixed and static, limiting its ability to capture the dynamic and evolving nature of personalization in real-world use.

In contrast, \ours introduces several design choices that address these gaps:
\begin{enumerate}
\item\textbf{Realistic Environment and App diversity.}
Unlike many benchmarks that rely on simulated environments or limited app support, \ours is deployed on physical smartphones and spans 22 third-party apps, covering diverse real-world scenarios such as e-commerce, food delivery, and travel. This setting ensures that agents are evaluated under realistic constraints (e.g., UI latency, pop-ups, network variability), increasing ecological validity and practical relevance.
\item\textbf{Dynamic Modeling of User Preferences.}
Most existing personalized benchmarks(e.g., FingerTip) either fix user preferences or omit them entirely. In contrast, \ours models both short-term and long-term user preferences to generate realistic, user-aligned task instructions. Our user profiles and behaviors are not static; instead, they evolve over time, enabling the study of adaptive personalization strategies that reflect real usage patterns.
\item\textbf{Multi-Dimensional Task Difficulty.}
While previous benchmarks typically define difficulty via task execution length, \ours introduces a richer difficulty model. We consider multiple dimensions, including both task execution length and instruction ambiguity. This enables a more nuanced evaluation of an agent's ability to reason about and adapt to varying personalized tasks.
\item\textbf{Fine-Grained and Incremental Personalized Evaluation Metrics.}
Existing evaluation metrics such as success signal or task completion ratio are often coarse-grained and overly reliant on fixed action sequences. \ours introduces process-oriented evaluation via task decomposition graphs, enabling step-by-step tracking of progress toward user-specific goals. Furthermore, we support incremental evaluation metrics to evaluate the performance changes of agents in long-term personalized use, which is missing in existing benchmarks.
\end{enumerate}
To summarize, \ours combines realistic environments, dynamic user modeling, and fine-grained evaluation to address key gaps in existing benchmarks for personalized GUI agents.

\section{Construction Task Decomposition Graph}
\label{sec:task_graph_construction}
We illustrate the constructed task decomposition graph with four personalized scenarios: shopping, dining, navigation, and travel. In the these task decomposition graphs, the nodes with green color refer to fixed type, while nodes with blue color refers to personalized type.

\begin{figure*}[h]
   \centering
   \includegraphics[width=0.75\linewidth]{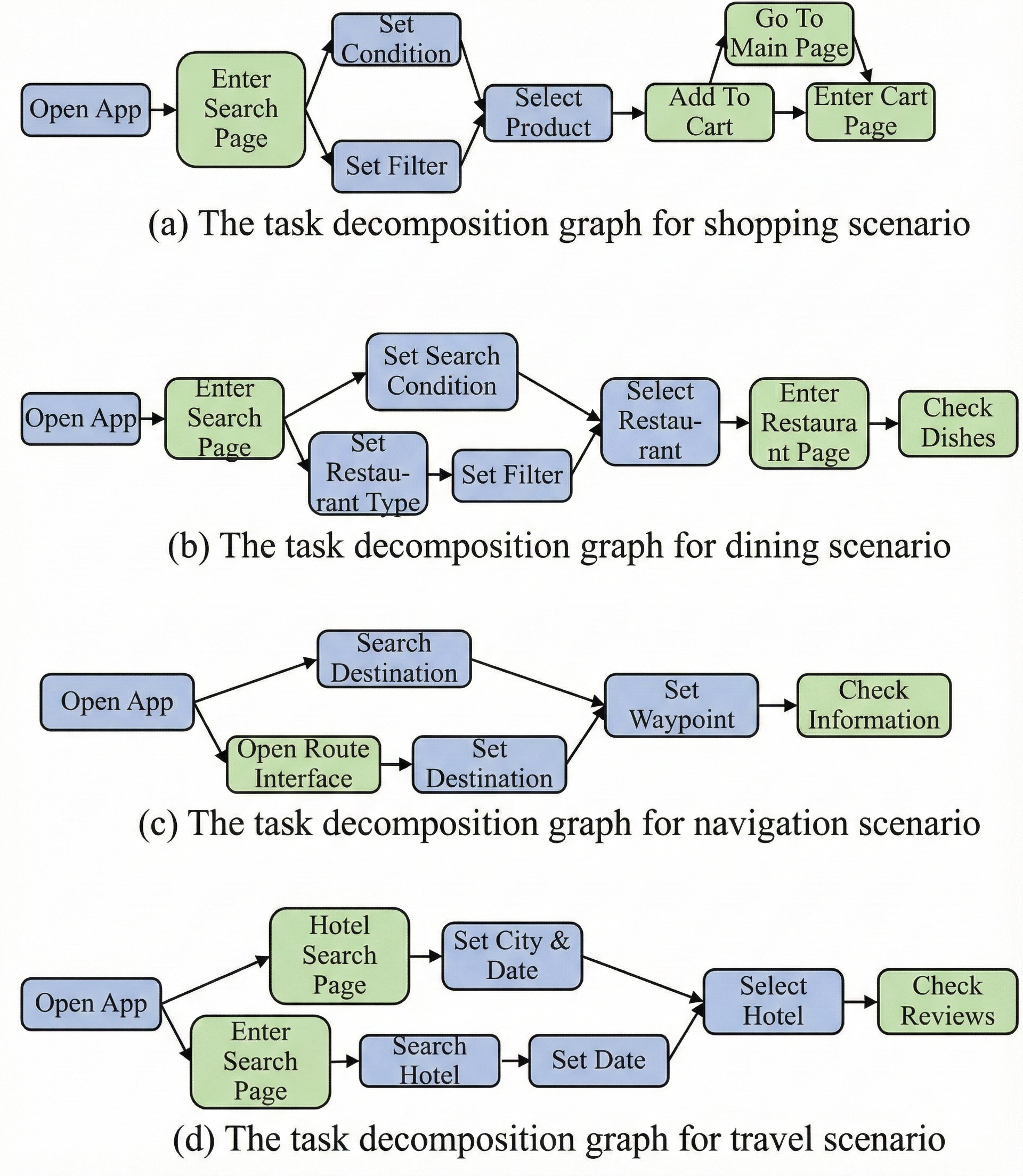}
   \caption{The task decomposition graphs of 4 personalized scenarios: shopping, dining, navigation, and travel. The nodes with green color refer to fixed type, while nodes with blue color refers to personalized type.
   } 
\label{fig:Shopping_graph}
\end{figure*}

\section{Instruction Instantiation}
\label{sec:instruction_instantiation}

Once a template is selected, we instantiate it by filling each slot with concrete values drawn from the user's preference distribution.  
Given a user profile $\rho$, which contains a set of preferences, each slot $l$ is filled by sampling a value $v_l \sim D_{\tau(l)}^{\rho}$, where $D_{\tau}^{\rho}$ is a distribution over admissible values conditioned on $\rho$. The instantiation function
$f(\mathsf{TP}_\mathcal{G},\{v_l\}_{l\in \mathcal{L}}) \mapsto I$
produces a concrete instruction $I$ by replacing each slot with the sampled value, while ensuring that the chosen values respect the execution dependencies in $\mathcal{G}$.
In this framework, We model user preferences as a combination of:
(i) \textit{Long-term preferences}, capturing stable patterns over time (e.g., favoring eco-friendly brands), with weight $w_{\text{long}}(t)$ modeled by a logarithmic growth function.
(ii) \textit{Short-term preferences}, capturing transient or contextual interests (e.g., recent searches or trends), with weight $w_{\text{short}}(t)$ modeled by a sinusoidal function.
At each time step $t$, we compute the probability of sampling from each component as:
$P_{\text{long}}(t) = w_{\text{long}}(t) / [w_{\text{long}}(t) + w_{\text{short}}(t)]$, 
and 
$P_{\text{short}}(t) = w_{\text{short}}(t) / [w_{\text{long}}(t) + w_{\text{short}}(t)]$, respectively.

These probabilities guide the instantiation process for each slot. For example, the \texttt{<product\_category>} slot might be filled with ``organic snacks'' based on long-term preferences, or ``Halloween costumes'' based on short-term seasonal trends.  
This dynamic sampling ensures that the personalized instructions reflect both enduring habits and recent behaviors, rather than being static or randomly constructed.
The detailed formulation of preference weights, along with experiments evaluating the quality of the generated instructions, is provided in Appendix \ref{sec:details_of_preference_weights_calculation} and \ref{sec:instruction_quality_evaluation}, respectively.

\subsection{Details of Preference Weights Calculation}
\label{sec:details_of_preference_weights_calculation}
We now describe the preference weights used to model long-term and short-term user patterns. These formulations balance tractability with behavioral plausibility, so that generated instructions reflect both stable needs and dynamic fluctuations.

\paragraph{Long-Term Preference weights.}

Long-term preferences capture stable, gradually accumulating needs that reset upon satisfaction (e.g., recurring but not continuous interests). Let
\begin{equation}
t \in \mathbb{Z}_{\geq 0}, \quad L \in \mathbb{Z}_{\geq 0}, \quad C \in \mathbb{Z}_{\geq 1},
\end{equation}
denote the current day, the last day of satisfaction, and the cycle length in days, respectively. With a base weight \(b \in (0,1]\) and maximum allowed weight \(W_{\max} \in (0,1]\), we define
\begin{equation}
\Delta(t) = \max\{0,\,t - L\}, \qquad \tau(t) = \Delta(t) \bmod C.
\end{equation}

The long-term preference weight is given by:
\begin{equation}
w_{\text{long}}(t) = \min\left\{ W_{\max}, \, b + \frac{\ln(1+\tau(t)) (1-b)}{\ln(1+C)} \right\}.
\end{equation}

The logarithmic form reflects the intuition that the *marginal increase* in preference weight diminishes over time: interest accumulates steadily but slows as it approaches saturation. Resetting \(\tau(t)\) upon satisfaction models the release of long-term need after fulfillment. The normalization by \(\ln(1+C)\) guarantees that the weight grows smoothly to a normalized scale within each cycle, while clamping ensures that weights remain bounded by \(W_{\max}\). This choice is particularly suitable for modeling durable interests such as dietary habits, weekly routines, or recurring educational needs, which build up gradually until satisfied.

\paragraph{Short-Term Preference weights.}

Short-term preferences capture immediate, oscillatory needs that fluctuate with context. Let
\begin{equation}
t \in \mathbb{Z}_{\geq 0}, \quad C \in \mathbb{Z}_{\geq 1}, \quad \phi \in \mathbb{R},
\end{equation}
denote the current day, the cycle length in days, and a phase offset, respectively. With bounds \(0 < w_{\min} < w_{\max} \leq 1\), we define:
\begin{equation}
w_{\text{short}}(t) = w_{\min} + (w_{\max} - w_{\min}) \cdot \frac{\sin\left(\frac{2\pi}{C} t + \phi\right) + 1}{2}.
\end{equation}

The sinusoidal structure is chosen for its natural ability to model periodic fluctuations. It ensures smooth transitions between high and low preference intensities, while the parameters \(w_{\min}\) and \(w_{\max}\) bound the oscillation range. This reflects short-term needs such as daily mood, time-of-day activity cycles, or context-specific demands (e.g., preferring entertainment in the evening versus productivity tools in the morning). Compared to stochastic fluctuations, sinusoidal oscillation avoids unnatural abrupt shifts and instead captures the rhythmic, recurrent nature of short-term preferences.

\begin{figure*}[t]
  \centering
  \includegraphics[width=1\linewidth]{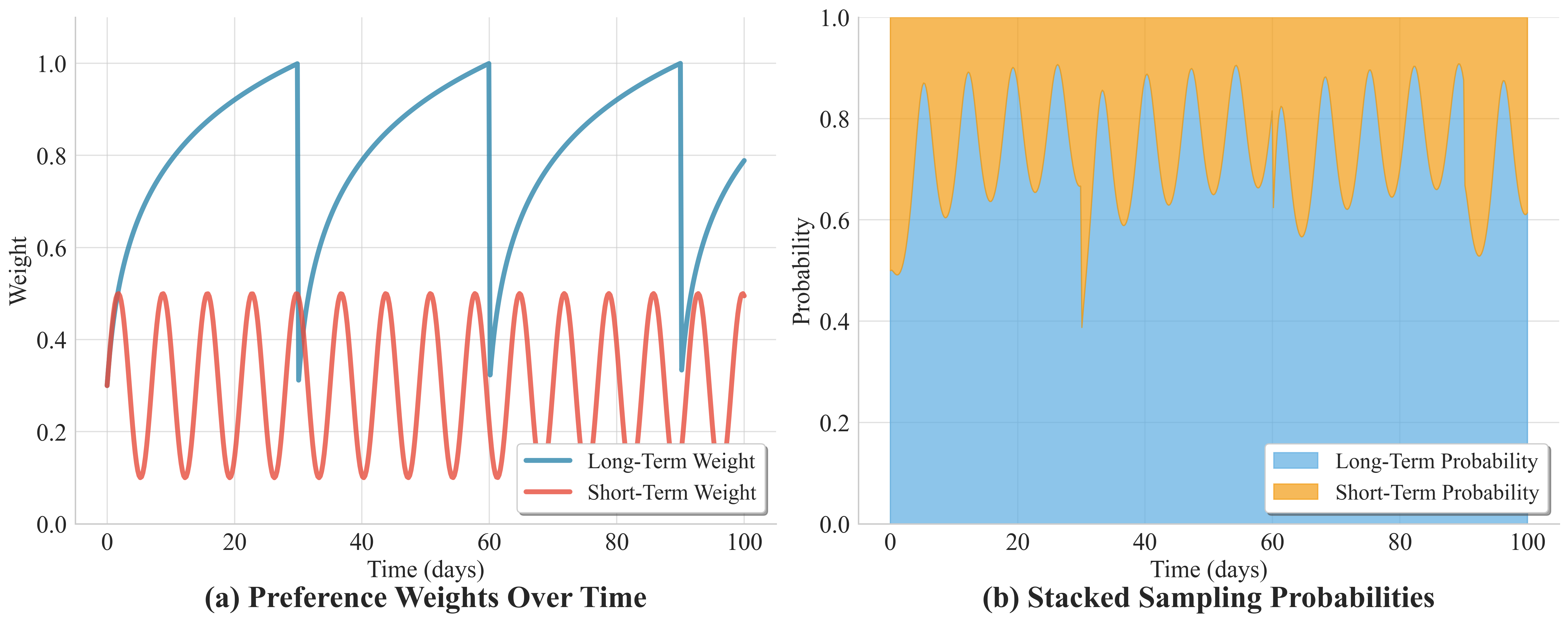}
  \caption{ (a) \textbf{Preference Weights Over Time}: The blue curve represents the long-term preference weight, which gradually increases with time, reflecting the accumulation of stable needs. The red curve depicts the short-term preference weight, which oscillates periodically, reflecting dynamic, context-dependent needs.  
  (b) \textbf{Stacked Sampling Probabilities}: The stacked area plot shows the normalized sampling probabilities over time. The blue region corresponds to the probability of sampling the long-term preference, while the orange region corresponds to the probability of sampling the short-term preference. The relative areas of these regions change over time, depending on the weights of each preference.
  } 
\label{fig:dual_weights_probabilities}
\end{figure*}

\paragraph{Normalized Sampling probabilities.}

Given both weights, we define the probabilities of drawing from long-term and short-term preferences as:
\begin{align}
P_{\text{long}}(t) = \frac{w_{\text{long}}(t)}{w_{\text{long}}(t) + w_{\text{short}}(t)}, \\
P_{\text{short}}(t) = \frac{w_{\text{short}}(t)}{w_{\text{long}}(t) + w_{\text{short}}(t)}.
\end{align}
This normalization yields a valid probability distribution and balances stability with variability: when long-term needs are strong, they dominate; when short-term fluctuations peak, they take precedence. A visual illustration of these dynamics can be found in Figure~\ref{fig:dual_weights_probabilities}.

\subsection{Instruction Quality Evaluation}
\label{sec:instruction_quality_evaluation}

To rigorously evaluate the quality of generated task instructions, we conducted a human evaluation study across \textbf{10 distinct scenarios} involving a total of \textbf{100 users}. For each user, we selected the \textbf{first 10 generated instructions} from each method and used them as the evaluation material. This design ensures a fair and comparable assessment across methods, while also reflecting the diversity of tasks encountered in practical settings.  

We compared three approaches for generating user instructions:  

\begin{enumerate}
    \item \textbf{Static Preference.}  
    Each user is assigned a \textit{fixed preference profile} that does not change over time. Instructions are generated by filling task templates strictly according to this static preference. While this captures general user tendencies, it cannot adapt to temporal shifts in user needs.  

    \item \textbf{Random Preference.}  
    At each time step, a \textit{random preference} is assigned to the user without considering historical data. Instructions are then generated from this random preference. Although this produces variety, the lack of temporal or contextual grounding leads to incoherent or irrelevant instructions.  

    \item \textbf{Dynamic Preference Modeling (Ours).}  
    We model user preferences as a distribution combining both \textit{long-term patterns} (stable, accumulated needs) and \textit{short-term patterns} (immediate, oscillating needs). At each time step $t$, the probability of sampling long-term versus short-term preferences is determined by their respective weights. Task instructions are generated based on the sampled preferences, enabling outputs that capture both the \textit{stable characteristics} of the user and their \textit{short-term dynamics}.
\end{enumerate}

Annotators were presented with \textbf{pairs of instruction sets} (10 instructions per set, per user). Each pair was evaluated \textbf{blindly}, comparing two methods at a time. Annotators judged which set was better overall, based on two criteria:  

\begin{itemize}
   \item \textbf{Plausibility / Realism:} Are the instructions natural and reasonable, resembling tasks that a real user might genuinely perform?  
   \item \textbf{Relevance to User Context:} Are the instructions aligned with the user's scenario, needs, and historical behavior?
\end{itemize}

Annotators were allowed to declare a \textbf{Tie} if the two sets were of equal quality. Each pair was evaluated by multiple annotators, and final results were aggregated using majority voting.  

\begin{table*}[h]
\caption{Pairwise win-rates of different methods in human evaluation.}
\label{tab:instruction_quality_evaluation}
\centering
\begin{tabular}{lccc}
\toprule
\textbf{Comparison (Pairwise)} & \textbf{Win-rate Ours $\uparrow$} & \textbf{Win-rate Baseline} & \textbf{Tie (\%)} \\
\midrule
Ours vs. Static   & 72.5\% & 24.1\% & 3.4\% \\
Ours vs. Random   & 81.3\% & 16.2\% & 2.5\% \\
Static vs. Random & 59.6\% & 38.7\% & 1.7\% \\
\bottomrule
\end{tabular}
\end{table*}

As shown in Table \ref{tab:instruction_quality_evaluation}, the results clearly demonstrate the advantage of \textbf{dynamic preference modeling}:  

\begin{itemize}
    \item \textbf{Ours vs. Static:} Our method achieved a significantly higher win-rate, confirming that accounting for temporal dynamics provides superior alignment with user preferences compared to fixed profiles.  
    \item \textbf{Ours vs. Random:} The large margin shows that random preferences fail to capture user needs, leading to inconsistent instruction quality.  
    \item \textbf{Static vs. Random:} While Static outperforms Random to some extent, it still lags far behind our method due to its inability to adapt to evolving preferences.  
\end{itemize}

The \textbf{low tie percentage} across all comparisons indicates that evaluators could generally distinguish between methods, supporting the reliability of the results.  

\section{Evaluation Details}
\label{sec:evaluation_details}
\begin{table*}[h]
  \centering
  \renewcommand{\arraystretch}{1.5}
  \footnotesize
  \caption{Evaluation metrics for the Personalized GUI Agent Task. 
  Short-term metrics measure absolute performance on individual tasks, 
  while long-term metrics capture improvements across instruction sequences. N denotes the number of tasks.}
  \label{tab:evaluation_metrics}
  \begin{tabular}{ccc}
  \toprule
  \textbf{Dimension} & \textbf{Immediate Objective} & \textbf{Long-term Objective} \\ 
  \midrule
  \multirow{2}{*}{Performance} 
    & APR: $\tfrac{1}{N}\sum \tfrac{\text{completed unit inst.}}{\text{total unit inst.}}$ 
    & $\Delta$APR: $\text{APR}_{T_k} - \text{APR}_{T_1}$ \\
    & PPR: $\tfrac{1}{N}\sum \tfrac{\text{completed flexible unit inst.}}{\text{total flexible unit inst.}}$ 
    & $\Delta$PPR: $\text{PPR}_{T_k} - \text{PPR}_{T_1}$ \\ \midrule
  \multirow{2}{*}{Efficiency} 
    & CT: $\tfrac{1}{N}\sum t_i$ 
    & $\Delta$CT: $\text{CT}_{T_k} - \text{CT}_{T_1}$ \\
    & CPT: $\tfrac{1}{N}\sum \text{Cost}_i$
    & $\Delta$CPT: $\text{CPT}_{T_k} - \text{CPT}_{T_1}$ \\ 
  \bottomrule
  \end{tabular}
\end{table*}

Table~\ref{tab:evaluation_metrics} organizes the evaluation metrics for the Personalized GUI Agent Task into two key dimensions---\textbf{performance} and \textbf{efficiency}---each assessed through both immediate and long-term objectives. Immediate metrics capture task-level outcomes, including the \textbf{A-process Ratio (APR)}, measuring the proportion of completed unit instructions, and the \textbf{P-process Ratio (PPR)}, restricted to flexible preference-sensitive nodes. Efficiency is evaluated via \textbf{Completion Time (CT)}, the average execution time per task, and \textbf{Cost per Task (CPT)}, the average token-based monetary cost. Long-term metrics (\(\Delta\)APR, \(\Delta\)PPR, \(\Delta\)CT, \(\Delta\)CPT) reflect improvements across task sequences, defined as the difference between performance at a later task \(T_k\) and the initial task \(T_1\). This framework jointly measures execution quality, personalization, time, and cost, providing a comprehensive view of both short-term performance and long-term adaptability.

\subsection{Metric Comparison}
\label{sec:metric_comparison}
\begin{figure*}[t]
   \centering
   \includegraphics[width=1\linewidth]{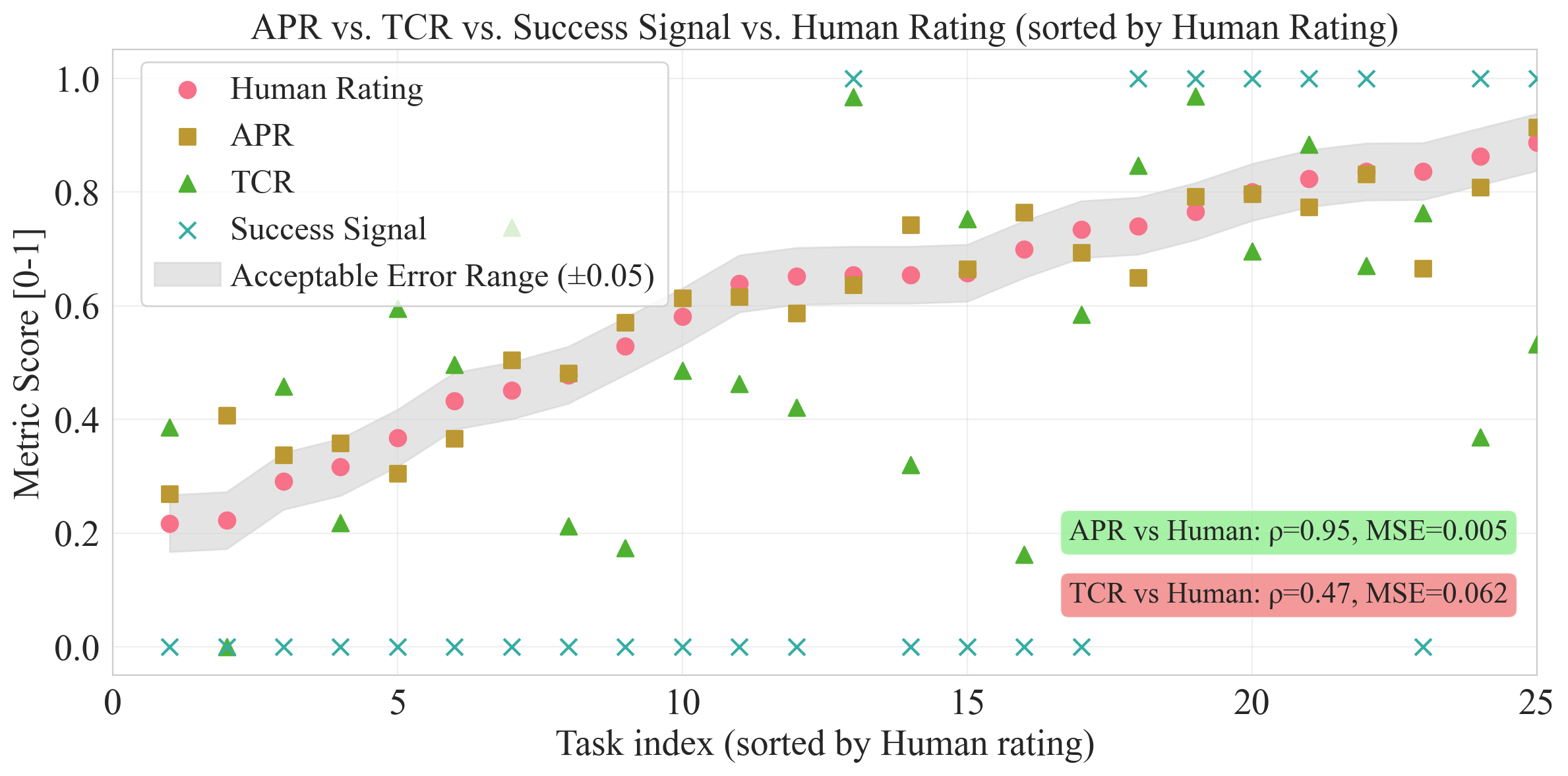}
   \caption{Comparison of evaluation metrics across tasks sorted by human ratings. Each point represents one task. Human ratings (pink dots) are shown with an acceptable tolerance band (±0.05, gray area). A-process Ratio (APR, brown squares) closely follows human ratings with minimal deviation, demonstrating strong alignment ($\rho$=0.95, MSE=0.005). In contrast, Task Completion Ratio (TCR, green triangles) shows weaker consistency with human ratings ($\rho$ =0.47, MSE=0.062), while Success Signal (blue crosses) provides only binary outcomes, lacking granularity.
   } 
\label{fig:metric_compare}
\end{figure*}

\textbf{Experiment Setup.} 
To evaluate the effectiveness of different metrics, we sampled 25 personalized GUI tasks from \ours and used \textbf{M3A} to execute each task and obtain the execution trajectory. For every task, we then measured the performance based on the trajectory using four metrics: \textbf{Success Signal}, \textbf{Task Completion Ratio (TCR)}, \textbf{A-process Ratio (APR)}, and \textbf{Human Ratings} (collected from third-party annotators). Success Signal provides a binary success/failure outcome, TCR measures the proportion of actions completed along a predefined reference path, APR reflects the average completion rate across unit instructions along the execution path, and annotators rate the completion ratio (human ratings) of each execution trajectory, which serves as the ground truth.

\textbf{Results and Analysis.} 
Figure~\ref{fig:metric_compare} compares the four metrics across tasks (sorted by human ratings). Success Signal is highly coarse-grained, producing only binary outcomes (0 or 1). This representation collapses diverse execution trajectories into two categories, obscuring important differences between agents that achieve partial progress versus those that fail entirely. TCR provides more variation but is still tightly coupled to a fixed path, which leads to penalizing valid alternative behaviors. As a result, TCR diverges notably from human ratings, with weaker correlation ($\rho=0.47$) and higher error (MSE=0.062).

By contrast, APR demonstrates strong consistency with human judgments. As shown in Figure~\ref{fig:metric_compare}, APR closely follows human ratings across all tasks, with almost all values lying within the human-acceptable tolerance band ($\pm 0.05$). Quantitatively, APR achieves a high correlation with human ratings ($\rho=0.95$) and near-zero error (MSE=0.005). These results show that APR captures fine-grained progress, tolerates diverse execution strategies, and aligns well with human evaluators. Compared to Success Signal and TCR, APR offers a more reliable assessment of agent performance in personalized, long-horizon GUI tasks.

\subsection{Metrics Calculation}
\label{sec:metrics_calculation}

While the metrics above provide a fine-grained view of agent behavior, computing them reliably is non-trivial. 
\citep{Xing2024} measure task completion by comparing executed actions against a fixed ground-truth sequence. 
Although useful, this method is tied to a single action path and cannot capture an agent's ability to follow alternative strategies or adapt to user preferences. 
To address this limitation, we leverage the TDG as an intermediate representation and convert it into a structured checklist. 
The checklist encodes execution dependencies, alternative paths, and the distinction between fixed and flexible unit instructions. 
We then employ a LLM such as GPT-5 as an evaluator: given the agent's execution trace and the checklist, the model determines which unit instructions were successfully completed. 
This graph-driven approach allows APR and PPR to be computed consistently, even when tasks admit multiple valid solutions or require personalization. 
To ensure the validity and reproducibility of this LLM-based evaluation, we conduct a comprehensive human validation study, as detailed in Section~\ref{sec:alignment_validation}.
Details of checklist template are provided in Appendix~\ref{sec:details_of_checklists_and_prompt_templates}.

\paragraph{Path Selection Algorithm.}
As described in Equation~\ref{eq:path_selection}, selecting the optimal path from the TDG is crucial for accurate metric computation. Algorithm~\ref{alg:path_selection} formalizes our two-stage alignment procedure:

\begin{algorithm}[h]
\caption{Trace-Graph Alignment with Path Selection}
\label{alg:path_selection}
\small
\begin{algorithmic}[1]
\REQUIRE Trace $\mathcal{T} = (a_1, \ldots, a_L)$, TDG $\mathcal{G}$ with valid paths $\mathcal{P}$
\ENSURE Optimal path $P^*$, alignment mapping $M$
\STATE \textbf{Stage 1: Action-to-Instruction Alignment}
\FOR{each action $a_i \in \mathcal{T}$}
    \STATE $m(a_i) \gets \text{LLM\_Align}(a_i, \mathcal{U})$ \COMMENT{Map action to unit instruction or $\varnothing$}
\ENDFOR
\STATE \textbf{Stage 2: Path Selection}
\STATE $\text{best\_score} \gets -1$; $\text{candidates} \gets \emptyset$
\FOR{each path $P \in \mathcal{P}$}
    \STATE $\text{score}(P) \gets |\{u \in P : \exists\, a_i, m(a_i) = u\}|$ \COMMENT{Count matched nodes}
    \IF{$\text{score}(P) > \text{best\_score}$}
        \STATE $\text{best\_score} \gets \text{score}(P)$; $\text{candidates} \gets \{P\}$
    \ELSIF{$\text{score}(P) = \text{best\_score}$}
        \STATE $\text{candidates} \gets \text{candidates} \cup \{P\}$
    \ENDIF
\ENDFOR
\STATE \textbf{Stage 3: Tie-Breaking} (if $|\text{candidates}| > 1$)
\STATE $P^* \gets \text{TieBreak}(\text{candidates})$ \COMMENT{Apply rules: flexible $\to$ length $\to$ earliest}
\RETURN $P^*$, $\{(a_i, m(a_i)) : m(a_i) \in P^*\}$
\end{algorithmic}
\end{algorithm}

The tie-breaking procedure (Lines 17--18) ensures deterministic path selection: we first prefer paths with more matched flexible nodes (prioritizing personalization), then shorter paths (favoring efficiency), and finally paths whose first matched node appears earliest in the trace (reflecting primary intent). This hierarchical criterion guarantees reproducible APR/PPR computation across different evaluation runs.

\subsection{Human Validation of LLM-based Alignment}
\label{sec:alignment_validation}

To validate the reliability of our LLM-based trace--graph alignment, we conduct a comprehensive human agreement study. 

\paragraph{Annotation Protocol.}
We randomly sampled 100 execution traces from a representative four-scenario subset of our benchmark (Shopping, Dining, Navigation, and Travel) with balanced complexity levels. Three trained annotators independently labeled each trace by marking which unit instructions in the corresponding TDG were successfully completed. Annotators were provided with the execution trace (screenshots and action logs), the TDG checklist, and detailed annotation guidelines. Prior to the main annotation, annotators completed a calibration phase on 20 pilot traces to ensure consistent interpretation of completion criteria.

\paragraph{Agreement Analysis.}
We measure inter-annotator agreement using Fleiss' Kappa ($\kappa$), which accounts for chance agreement among multiple raters. The overall inter-annotator agreement reached $\kappa=0.87$, indicating \textit{almost perfect agreement} according to standard interpretation guidelines \citep{Landis1977}. For cases with disagreement, we resolved conflicts through majority voting to establish the ground-truth labels.

\paragraph{LLM Alignment Accuracy.}
We then compared the LLM-based alignment results against the human-annotated ground truth. Table~\ref{tab:alignment_validation} summarizes the results across different instruction types:

\begin{table}[h]
\centering
\caption{Validation of LLM-based alignment against human annotations. We report Accuracy, Precision, Recall, F1-score, and Cohen's $\kappa$ for agreement with human ground truth.}
\label{tab:alignment_validation}
\small
\begin{tabular}{@{}lccccc@{}}
\toprule
\textbf{Instruction Type} & \textbf{Acc.} & \textbf{Prec.} & \textbf{Rec.} & \textbf{F1} & \textbf{$\kappa$} \\ \midrule
Fixed Instructions   & 0.94 & 0.93 & 0.96 & 0.94 & 0.88 \\
Flexible Instructions & 0.91 & 0.90 & 0.93 & 0.91 & 0.83 \\
\midrule
\textbf{Overall}      & 0.93 & 0.92 & 0.95 & 0.93 & 0.86 \\ \bottomrule
\end{tabular}
\end{table}

The LLM-based alignment achieves an overall accuracy of 93\% and Cohen's $\kappa=0.86$ with human annotations, demonstrating strong agreement. Fixed instructions show slightly higher accuracy (94\%) than flexible instructions (91\%), which is expected given that flexible instructions involve more nuanced judgments about user preference satisfaction.

\paragraph{Reproducibility Analysis.}
To assess the reproducibility of the LLM-based alignment, we ran the evaluation pipeline five times on the same set of 100 traces using identical prompts and model settings (temperature = 0). The alignment results showed 98.7\% consistency across runs, with a standard deviation of only 0.8\% for the computed APR values. This high consistency demonstrates that our LLM-based alignment method is reproducible when using deterministic inference settings.

\paragraph{Error Analysis.}
We analyzed the 7\% of cases where LLM alignment disagreed with human annotations. The primary sources of disagreement were: (i) ambiguous partial completions where an instruction was partially but not fully satisfied (42\% of errors), (ii) alternative valid paths not explicitly enumerated in the checklist (31\%), and (iii) edge cases involving UI state changes between screenshots (27\%). These insights inform future improvements to both the checklist design and the evaluation prompts.

\subsection{Details of Checklists and Prompt Templates}
\label{sec:details_of_checklists_and_prompt_templates}
\paragraph{Checklist Structure.} 
We provide four domain-specific checklist templates corresponding to a representative four-scenario subset of our benchmark: \textbf{Shopping}, \textbf{Dining}, \textbf{Navigation}, and \textbf{Travel}. Each checklist is derived from the Task Decomposition Graph (TDG) and contains the following structured elements for each unit instruction:

\begin{itemize}[leftmargin=*, labelsep=5pt]
    \item \textbf{Checkbox}: A visual marker ($\square$) indicating completion status.
    \item \textbf{Step Name}: The name of the action, annotated with its operation type (\textit{Fixed} or \textit{Flexible}).
    \item \textbf{Description}: A clear explanation of what the step accomplishes.
    \item \textbf{Dependency}: The prerequisite step(s) that must be completed before this step can be executed.
    \item \textbf{Path Options}: For flexible steps with multiple valid execution paths, we enumerate them as Path A, Path B, etc., each with its own sub-steps.
\end{itemize}

The distinction between \textit{Fixed} and \textit{Flexible} operations is important: Fixed steps are mandatory with no alternatives, while Flexible steps allow multiple valid approaches or are influenced by user preferences. This design lets the checklist accommodate personalized execution strategies while keeping evaluation criteria consistent.

\paragraph{Prompt Templates.} We design two prompts to facilitate checklist-based evaluation:
\begin{itemize}[leftmargin=*, labelsep=5pt]
    \item \textbf{Prompt 1} (Checklist Generation): Converts a visual Task Decomposition Graph into a structured textual checklist that can be processed by LLMs. The prompt instructs the model to extract step names, descriptions, dependencies, and operation types while preserving the graph's hierarchical structure.
    \item \textbf{Prompt 2} (LLM-based Evaluation): Given an agent's execution trace (as a sequence of screenshots) and the corresponding checklist, the evaluator LLM determines which unit instructions were successfully completed. The prompt specifies evaluation rules, including handling of alternative paths and dependency constraints.
\end{itemize}

The complete checklist templates for all four scenarios and the detailed prompt templates are presented in Checklist 1--4 and Prompt 1--2.

\section{Templates of personalized GUI instruction}
\label{sec:templates_of_personalized_gui_instruction}
We provide examples of templates in personalized scenarios. The templates are categorized by \textit{clarity level} (high, medium, low) and \textit{complexity level} (low, medium, high). In each level, we provide 3 templates with different expressions, which improve the variety of personalized instructions.

\section{LLM Usage}
Large Language Models (LLMs) were used to aid in the writing and polishing of the manuscript. Specifically, we used an LLM to assist in refining the language, improving readability, and ensuring clarity in various sections of the paper. The model helped with tasks such as sentence rephrasing, grammar checking, and enhancing the overall flow of the text.

The authors take full responsibility for the content of the manuscript, including any text generated or polished by the LLM. We have ensured that the LLM-generated text adheres to ethical guidelines and does not contribute to plagiarism or scientific misconduct.

\begin{figure*}
\begin{tcolorbox}[colback=blue!10!white, colframe=blue!70!black, title=\textbf{Checklist 1: Shopping sceneario}, fontupper=\small, after skip=4pt]
\begin{quote}
\begin{itemize}[leftmargin=*, labelsep=5pt]
  \item[\textcolor{gray}{$\square$}] \textbf{Open Taobao Fixed} \\
    \textit{Description:} Open the Taobao homepage and enter the main platform page
    
  \item[\textcolor{gray}{$\square$}] \textbf{Enter Search Page Fixed} \\
    \textit{Description:} Enter the search interface to prepare for product search \\
    \textit{Dependency:} Must be executed after ``Open Taobao''
    
  \item[\textcolor{gray}{$\square$}] \textbf{Set Conditions (choose one path)} \\
    \textit{Description:} Set search constraints (either method is acceptable)
    \textit{Dependency:} Must first complete ``Enter Search Page''
    \begin{itemize}[leftmargin=*, labelsep=5pt]
      \item \textbf{Path A:}
        \begin{itemize}[leftmargin=*, labelsep=5pt]
          \item[\textcolor{gray}{$\square$}] \textbf{Set Search Conditions Flexible} \\
            \textit{Description:} Enter keywords, categories, etc. for searching
        \end{itemize}
      \item \textbf{Path B:}
        \begin{itemize}[leftmargin=*, labelsep=5pt]
          \item[\textcolor{gray}{$\square$}] \textbf{Set Filter Conditions Flexible} \\
            \textit{Description:} Set filters such as price range, sales sorting, etc.
        \end{itemize}
    \end{itemize}
  \item[\textcolor{gray}{$\square$}] \textbf{Select Product Flexible} \\
    \textit{Description:} Select a product from search results \\
    \textit{Dependency:} Must first complete ``Set Conditions'' in either path
  \item[\textcolor{gray}{$\square$}] \textbf{Add to Cart Fixed} \\
    \textit{Description:} Click the ``Add to Cart'' button \\
    \textit{Dependency:} Must first complete ``Select Product''
  \item[\textcolor{gray}{$\square$}] \textbf{Navigate to Cart Page (choose one path) Flexible} \\
    \textit{Description:} Enter the shopping cart page (flexible path)
    \textit{Dependency:} Must first complete ``Add To Cart''
    \begin{itemize}[leftmargin=*, labelsep=5pt]
      \item \textbf{Path A:}
        \begin{itemize}[leftmargin=*, labelsep=5pt]
          \item[\textcolor{gray}{$\square$}] \textbf{Enter Cart Page (directly) Fixed} \\
            \textit{Description:} Directly enter the shopping cart page after adding to cart
        \end{itemize}
      \item \textbf{Path B:}
        \begin{itemize}[leftmargin=*, labelsep=5pt]
          \item[\textcolor{gray}{$\square$}] \textbf{Return to Home Page Fixed} \\
            \textit{Description:} First return to the homepage
          \item[\textcolor{gray}{$\square$}] \textbf{Enter Cart Page (from Home) Fixed} \\
            \textit{Description:} Enter the shopping cart page from the homepage \\
            \textit{Dependency:} Must first complete ``Return to Home Page''
        \end{itemize}
    \end{itemize}
\end{itemize}
\end{quote}
\end{tcolorbox}
\end{figure*}
\begin{figure*}
\begin{tcolorbox}[colback=blue!10!white, colframe=blue!70!black, title=\textbf{Checklist 2: Dining sceneario}, fontupper=\small, before skip=4pt, after skip=4pt]
\begin{quote}
\begin{itemize}[leftmargin=*, labelsep=5pt]

  \item[\textcolor{gray}{$\square$}] \textbf{Open App Flexible} \\
    \textit{Description:} Open the app and enter its main page
  \item[\textcolor{gray}{$\square$}] \textbf{Enter Search Page Fixed} \\
    \textit{Description:} Enter the search interface to prepare for product search \\
    \textit{Dependency:} Must be executed after ``Open App''

  \item[\textcolor{gray}{$\square$}] \textbf{Set Conditions (choose one path) Flexible} \\
    \textit{Description:} Set Searching Conditions (flexible path)\\
    \textit{Dependency:} Must be executed after ``Enter Search Page''
    \begin{itemize}[leftmargin=*, labelsep=5pt]
      \item \textbf{Path A:}
        \begin{itemize}[leftmargin=*, labelsep=5pt]
          \item[\textcolor{gray}{$\square$}] \textbf{Set Search Conditions Flexible} \\
            \textit{Description:} Directly search for the restaurant using keywords
        \end{itemize}
        
      \item \textbf{Path B:}
        \begin{itemize}[leftmargin=*, labelsep=5pt]
          \item[\textcolor{gray}{$\square$}] \textbf{Set Restaurant Type Flexible} \\
            \textit{Description:} Search for the type or name of the restaurant
            
          \item[\textcolor{gray}{$\square$}] \textbf{Set Filter Conditions Flexible} \\
            \textit{Description:} Set filter conditions such as price, distance,etc. \\
            \textit{Dependency:} Must first complete ``Set Restaurant Type''
        \end{itemize}
    \end{itemize}

  \item[\textcolor{gray}{$\square$}] \textbf{Select Restaurant Flexible} \\
    \textit{Description:} Select a restaurant from the searching results \\
    \textit{Dependency:} Must be executed after ``Set Searching Conditions''
    
  \item[\textcolor{gray}{$\square$}] \textbf{Enter Main Page of Restaurant Fixed} \\
    \textit{Description:} Click to enter the main page of the restaurant \\
    \textit{Dependency:} Must be executed after ``Select Restaurant''

  \item[\textcolor{gray}{$\square$}] \textbf{Check Recommended Dishes Fixed} \\
    \textit{Description:} Check the restaurant's recommended dishes \\
    \textit{Dependency:} Must be executed after ``Enter Main Page of Restaurant''
\end{itemize}
\end{quote}
\end{tcolorbox}
\end{figure*}
\begin{figure*}
\begin{tcolorbox}[colback=blue!10!white, colframe=blue!70!black, title=\textbf{Checklist 3: Navigation sceneario}, fontupper=\small, before skip=4pt, after skip=4pt]
\begin{quote}
\begin{itemize}[leftmargin=*, labelsep=5pt]
  \item[\textcolor{gray}{$\square$}] \textbf{Open App Flexible} \\
    \textit{Description:} Open the app and enter its main page

  \item[\textcolor{gray}{$\square$}] \textbf{Set Route To Destination Flexible} \\
    \textit{Description:}Set the navigation destination (flexible path)\\
    \textit{Dependency:} Must be executed after ``Open App''
    \begin{itemize}[leftmargin=*, labelsep=5pt]
      \item \textbf{Path A:}
        \begin{itemize}[leftmargin=*, labelsep=5pt]
          \item[\textcolor{gray}{$\square$}] \textbf{Search For Destination Flexible} \\
            \textit{Description:} Directly search for the destination in the searching box
        \end{itemize}
      \item \textbf{Path B:}
        \begin{itemize}[leftmargin=*, labelsep=5pt]
          \item[\textcolor{gray}{$\square$}] \textbf{Open Route Planning Interface Fixed} \\
            \textit{Description:} Enter the route planning and destination searching interface
            
          \item[\textcolor{gray}{$\square$}] \textbf{Set Destination Flexible} \\
            \textit{Description:} Set destination in the route planning interface \\
            \textit{Dependency:} Must first complete ``Open Route Planning Interface''
        \end{itemize}
    \end{itemize}
    
  \item[\textcolor{gray}{$\square$}] \textbf{Set Waypoint Flexible} \\
    \textit{Description:} Set the waypoint for the route \\
    \textit{Dependency:} Must be executed after ``Set Route To Destination''

  \item[\textcolor{gray}{$\square$}] \textbf{Check Route Information Fixed} \\
    \textit{Description:} Check the information of the route \\
    \textit{Dependency:} Must be executed after ``Set Waypoint''
\end{itemize}
\end{quote}
\end{tcolorbox}
\end{figure*}
\begin{figure*}
\begin{tcolorbox}[colback=blue!10!white, colframe=blue!70!black, title=\textbf{Checklist 4: Travel sceneario}, fontupper=\small, before skip=4pt, after skip=4pt]
\begin{quote}
\begin{itemize}[leftmargin=*, labelsep=5pt]
  \item[\textcolor{gray}{$\square$}] \textbf{Open App Flexible} \\
    \textit{Description:} Open the App homepage and enter the main platform page
    
  \item[\textcolor{gray}{$\square$}] \textbf{Search For Hotel (choose one path)} \\
    \textit{Description:} Set search constraints (either method is acceptable)\
    \textit{Dependency:} Must first complete ``Open App''
    \begin{itemize}[leftmargin=*, labelsep=5pt]
      \item \textbf{Path A:}
        \begin{itemize}[leftmargin=*, labelsep=5pt]
          \item[\textcolor{gray}{$\square$}] \textbf{Enter Hotel Search Page Fixed} \\
            \textit{Description:} Enter the search category for hotel
        \end{itemize}
        \begin{itemize}[leftmargin=*, labelsep=5pt]
          \item[\textcolor{gray}{$\square$}] \textbf{Set City And Date Flexible} \\
            \textit{Description:} Set the searching conditions like city,type, and time
            \textit{Dependency:} Must first complete ``Enter Hotel Search Page''
        \end{itemize}
        
      \item \textbf{Path B:}
        \begin{itemize}[leftmargin=*, labelsep=5pt]
          \item[\textcolor{gray}{$\square$}] \textbf{Enter Search Page Fixed} \\
            \textit{Description:}Enter the search page of the APP
        \end{itemize}
        \begin{itemize}[leftmargin=*, labelsep=5pt]
          \item[\textcolor{gray}{$\square$}] \textbf{Search For Hotel In Target City Flexible} \\
            \textit{Description:} Set searching conditions like city and hotel type
            \textit{Dependency:} Must first complete ``Enter Search Page''
        \end{itemize}
        \begin{itemize}[leftmargin=*, labelsep=5pt]
          \item[\textcolor{gray}{$\square$}] \textbf{Set Date Fixed} \\
            \textit{Description:} Set dates of the stay
            \textit{Dependency:} Must first complete ``Set Date''
        \end{itemize}
    \end{itemize}
    
  \item[\textcolor{gray}{$\square$}] \textbf{Select Hotel Flexible} \\
    \textit{Description:} Select a hotel from the results
    \textit{Dependency:} Must first complete ``Search For Hotel''

  \item[\textcolor{gray}{$\square$}] \textbf{Check Reviews Fixed} \\
    \textit{Description:} Check reviews of the hotel on the APP
    \textit{Dependency:} Must first complete ``Select Hotel''
\end{itemize}
\end{quote}
\end{tcolorbox}
\end{figure*}
\clearpage
\label{sec:prompt_templates}

\begin{figure*}[t]
\begin{tcolorbox}[colback=gray!15, colframe=black!30, title=\textbf{Prompt 1: Prompt for turning task decomposition graphs into checklist:}, fontupper=\small]
\begin{quote}
    Your task is to turn a task decomposition graph in a picture into textual checklist that can be understood by another large language model easily. The task decomposition graph is provided below:\\
    \{TASK\_GRAPH\}\\
    The checklist you generate should include:\\ \\
    -\*\*A checkbox\*\*\\
    -\*\*Step Name\*\*: The name of the action from the graph.\\
    -\*\*Description \*\*:A clear explanation of what the step does.\\
    -\*\*Dependency \*\*:The prerequisite step(s) that must be completed before this step.\\
    -\*\*Operation Type\*\*: Mark as Fixed if the step is mandatory with no alternatives, or Flexible if there are multiple paths/options, or the step is affected by user preference. (including grouped parallel steps).\\
    -\*\*Path Options (if applicable)\*\*: If the step has multiple paths (e.g., parallel steps leading to the same next step), list them as Path A, Path B, etc., with a description with the same format as the steps you generate.\\ \\
    A example of the checklist you generate is provided below, make sure you follow the correct format.\\
    Example:
\begin{itemize}
    \item[\textcolor{gray}{$\square$}] \textbf{Set Route To Destination Flexible} \\
    \textit{Description:}Set the navigation destination (flexible path)\\
    \textit{Dependency:} Must be executed after ``Open App''
    \begin{itemize}[leftmargin=*, labelsep=5pt]
      \item \textbf{Path A:}
        \begin{itemize}[leftmargin=*, labelsep=5pt]
          \item[\textcolor{gray}{$\square$}] \textbf{Search For Destination Flexible} \\
            \textit{Description:} Directly search for the destination in the searching box
        \end{itemize}
        
      \item \textbf{Path B:}
        \begin{itemize}[leftmargin=*, labelsep=5pt]
          \item[\textcolor{gray}{$\square$}] \textbf{Open Route Planning Interface Fixed} \\
            \textit{Description:} Enter the route planning and destination searching interface
            
          \item[\textcolor{gray}{$\square$}] \textbf{Set Destination Flexible} \\
            \textit{Description:} Set destination in the route planning interface \\
            \textit{Dependency:} Must first complete ``Open Route Planning Interface''
        \end{itemize}
    \end{itemize}
\end{itemize}
\end{quote}
\end{tcolorbox}
\end{figure*}

\begin{figure*}[t]
\begin{tcolorbox}[colback=gray!15, colframe=black!30, title=\textbf{Prompt 2: Prompt for LLM-based evaluation:}, fontupper=\small]
\begin{quote}
    You are an expert in evaluating the performance of a GUI Agent. Based on the input checklist and the agent's execution flow images, you will determine one by one whether the agent has completed all sub-tasks in the checklist. The checklist will be provided to you in JSON format, while the images will be base64 encoded. Your output should also be a JSON, where you set the complete marker for each sub-task in the input JSON to 1 if the sub-task is completed and 0 if not completed.
    
    \{IMAGES\}\\
    \{CHECKLIST\}\\
    
    Rules:
    1. Check the subtasks and determine whether they are completed one by one.
    2. For subtask with various paths, the completion of any path marks the completion of the subtask.
    3. If the dependency of a subtask is not completed, the subtask itself is not completed as well.
    
\end{quote}
\end{tcolorbox}
\end{figure*}

\clearpage

\begin{table*}[t]
\centering
\caption{Instruction templates of shopping scenario.}
\renewcommand{\arraystretch}{1.1}
\begin{tabularx}{\textwidth}{|c|c|X|}
\hline
\textbf{Clarity level} & \textbf{Complexity level} & \textbf{Template Examples (English)} \\ \hline

\multirow{3}{*}{High clarity} 
& Low & 
1. I want to search for \{brand\} \{category\} on \{APP\}, with prices between \{price range\} yuan. \newline
2. Help me find a \{brand\} \{category\} on \{APP\}, with prices in the range of \{price range\} yuan. \newline
3. Check \{APP\} for a \{brand\} \{category\}, keeping the price within \{price range\} yuan. \\ \cline{2-3}

& Medium & 
1. I want to buy a \{brand\} \{category\} on \{APP\}, with prices between \{price range\} yuan, and stop at the product details page. \newline
2. Help me find a \{brand\} \{category\} on \{APP\}, with prices in the range of \{price range\} yuan, and stay on the product details page. \newline
3. Check \{APP\} for a \{brand\} \{category\}, with prices around \{price range\} yuan, and stay on the product details page. \\ \cline{2-3}

& High & 
1. I want to buy a \{brand\} \{category\} on \{APP\}, with prices between \{price range\} yuan, add the item to the cart, and finally stay on the cart page. \newline
2. Help me find a \{brand\} \{category\} on \{APP\}, with prices around \{price range\} yuan, add it to the cart, and finally stay on the cart page. \newline
3. Check \{APP\} for a \{brand\} \{category\}, within \{price range\} yuan, add it to the cart, and finally stay on the cart page. \\ \hline

\multirow{3}{*}{Medium clarity} 
& Low & 
1. Search \{brand\} \{category\} on \{APP\}, price about \{price range\}. \newline
2. Help me find \{brand\} \{category\} in \{APP\}, budget \{price range\}. \newline
3. Check \{APP\} for \{brand\} \{category\}, price around \{price range\}. \\ \cline{2-3}

& Medium & 
1. Want to buy \{brand\} \{category\} on \{APP\}, price \{price range\}, just open the details page. \newline
2. \{APP\}, find \{brand\} \{category\}, around \{price range\}, go to the details page. \newline
3. Help me search \{brand\} \{category\} on \{APP\}, price \{price range\}, open the details page. \\ \cline{2-3}

& High & 
1. Buy \{brand\} \{category\} on \{APP\}, budget \{price range\}, add to cart and stay there. \newline
2. \{APP\}, find \{brand\} \{category\}, price about \{price range\}, add to cart. \newline
3. Help me add \{brand\} \{category\} (\{price range\}) in \{APP\} to the cart, and stay on the cart page. \\ \hline

\multirow{3}{*}{Low clarity} 
& Low & 
1. \{APP\} \{brand\}\{category\} \{price range\} \newline
2. Find \{brand\}\{category\} \{price range\} @\{APP\} \newline
3. \{brand\}\{category\} \{price range\} /\{APP\} \\ \cline{2-3}

& Medium & 
1. \{APP\} \{brand\}\{category\} \{price range\} details \newline
2. \{brand\}\{category\} \{price range\} → details @\{APP\} \newline
3. \{APP\}: \{brand\}\{category\} \{price range\} details page \\ \cline{2-3}

& High & 
1. \{APP\} \{brand\}\{category\} \{price range\} cart \newline
2. Add to cart \{brand\}\{category\} \{price range\} @\{APP\} \newline
3. \{brand\}\{category\} \{price range\} → cart /\{APP\} \\ \hline

\end{tabularx}
\label{tab:templates_of_shopping_scenario}
\end{table*}

\begin{table*}[t]
\centering
\caption{Instruction templates for music scenario.}
\label{tab:templates_of_music_scenario}
\renewcommand{\arraystretch}{1.1}
\begin{tabularx}{\textwidth}{|c|c|X|}
\hline
\textbf{Clarity level} & \textbf{Complexity level} & \textbf{Template Examples (English)} \\ \hline

\multirow{3}{*}{High clarity} 
& Low & 
1. I want to search for \{singer\} on \{APP\} and go to the singer’s homepage. \newline
2. Help me search for \{singer\} on \{APP\} and enter the singer’s homepage. \newline
3. Check \{APP\} for \{singer\} and click to open the singer’s homepage. \\ \cline{2-3}

& Medium & 
1. I want to search for \{singer\} on \{APP\}, go to the singer’s homepage, and view the albums. \newline
2. Help me find \{singer\} on \{APP\}, enter the singer’s homepage, and check the albums. \newline
3. On \{APP\}, search for \{singer\}, open the singer’s homepage, then view the albums. \\ \cline{2-3}

& High & 
1. I want to search for \{singer\} on \{APP\}, open the singer’s homepage, check the albums, and select a song to play. \newline
2. Help me find \{singer\} on \{APP\}, enter the singer’s homepage, view the albums, and then play a song. \newline
3. On \{APP\}, search for \{singer\}, open the homepage, view the albums, and finally play a song. \\ \hline

\multirow{3}{*}{Medium clarity} 
& Low & 
1. Search \{singer\} on \{APP\}, go to homepage. \newline
2. Help me find \{singer\} in \{APP\}, check homepage. \newline
3. \{APP\} search \{singer\}, open homepage. \\ \cline{2-3}

& Medium & 
1. Want to find \{singer\} on \{APP\}, check homepage and albums. \newline
2. \{APP\} search \{singer\}, enter homepage, view albums. \newline
3. Help me search for \{singer\} on \{APP\}, view albums in homepage. \\ \cline{2-3}

& High & 
1. Find \{singer\} on \{APP\}, view albums, select a song to play. \newline
2. \{APP\} search \{singer\}, choose a song from album to listen. \newline
3. Help me find \{singer\} on \{APP\}, select a song from album and play. \\ \hline

\multirow{3}{*}{Low clarity} 
& Low & 
1. \{APP\} \{singer\} homepage \newline
2. Find \{singer\} @\{APP\} \newline
3. \{singer\} homepage /\{APP\} \\ \cline{2-3}

& Medium & 
1. \{APP\} \{singer\} album \newline
2. \{singer\} album @\{APP\} \newline
3. \{APP\}: \{singer\} album page \\ \cline{2-3}

& High & 
1. \{APP\} \{singer\} play album song \newline
2. Play \{singer\} album song @\{APP\} \newline
3. \{singer\} → play album song /\{APP\} \\ \hline

\end{tabularx}
\end{table*}

\begin{table*}[t]
  \centering
  \caption{Instruction templates for dining scenario.}
  \label{tab:templates_of_dining_scenario}
  \renewcommand{\arraystretch}{1.1}
  \begin{tabularx}{\textwidth}{|c|c|X|}
  \hline
  \textbf{Clarity level} & \textbf{Complexity level} & \textbf{Template Examples (English)} \\ \hline
  
  \multirow{3}{*}{High clarity} 
  & Low & 
  1. I want to search for \{cuisine type\} restaurants on \{APP\}, with an average spending between \{price range\} yuan. \newline
  2. Help me find a \{cuisine type\} restaurant on \{APP\}, with average spending in the range of \{price range\} yuan. \newline
  3. Check \{APP\} for \{cuisine type\} restaurants, priced between \{price range\} yuan. \\ \cline{2-3}
  
  & Medium & 
  1. I want to search for \{cuisine type\} restaurants on \{APP\}, with an average spending between \{price range\} yuan, and stay on the restaurant details page. \newline
  2. Help me find a \{cuisine type\} restaurant on \{APP\}, with average spending in the range of \{price range\} yuan, and remain on the details page. \newline
  3. Check \{APP\} for \{cuisine type\} restaurants, priced between \{price range\} yuan, and stay on the details page. \\ \cline{2-3}
  
  & High & 
  1. I want to search for \{cuisine type\} restaurants on \{APP\}, with an average spending between \{price range\} yuan, enter the details page, and view all recommended dishes. \newline
  2. Help me find a \{cuisine type\} restaurant on \{APP\}, with average spending in the range of \{price range\} yuan, go to the details page, and view all recommended dishes. \newline
  3. Check \{APP\} for \{cuisine type\} restaurants, priced between \{price range\} yuan, enter the details page, and view all recommended dishes. \\ \hline
  
  \multirow{3}{*}{Medium clarity} 
  & Low & 
  1. Search \{cuisine type\} on \{APP\}, around \{price range\}. \newline
  2. Help me find \{cuisine type\} in \{APP\}, budget \{price range\}. \newline
  3. Check \{APP\} for \{cuisine type\}, price about \{price range\}. \\ \cline{2-3}
  
  & Medium & 
  1. Want to find \{cuisine type\} on \{APP\}, price between \{price range\}, open details page. \newline
  2. \{APP\} search \{cuisine type\}, in range \{price range\}, open details. \newline
  3. Help me search \{cuisine type\} on \{APP\}, price \{price range\}, view details. \\ \cline{2-3}
  
  & High & 
  1. Find \{cuisine type\} on \{APP\}, budget \{price range\}, check recommended dishes. \newline
  2. \{APP\} search \{cuisine type\}, price \{price range\}, view recommended dishes. \newline
  3. Help me find \{cuisine type\} (\{price range\}) on \{APP\}, view recommended dishes. \\ \hline
  
  \multirow{3}{*}{Low clarity} 
  & Low & 
  1. \{APP\} \{cuisine type\} \{price range\} \newline
  2. Find \{cuisine type\} \{price range\} @\{APP\} \newline
  3. \{cuisine type\} \{price range\} /\{APP\} \\ \cline{2-3}
  
  & Medium & 
  1. \{APP\} \{cuisine type\} \{price range\} details \newline
  2. \{cuisine type\} \{price range\} → details @\{APP\} \newline
  3. \{APP\}: \{cuisine type\} \{price range\} details page \\ \cline{2-3}
  
  & High & 
  1. \{APP\} \{cuisine type\} \{price range\} view recommended dishes \newline
  2. Recommended dishes \{cuisine type\} \{price range\} @\{APP\} \newline
  3. \{cuisine type\} \{price range\} → view recommended dishes /\{APP\} \\ \hline
  
  \end{tabularx}
  \label{tab:templates_of_restaurant_search_scenario}
  \end{table*}





\end{document}

%% file: math_commands.tex
\usepackage{amsmath,amsfonts,bm}

\def\eqref#1{equation~\ref{#1}}

\def\1{\bm{1}}

\DeclareMathAlphabet{\mathsfit}{\encodingdefault}{\sfdefault}{m}{sl}
\SetMathAlphabet{\mathsfit}{bold}{\encodingdefault}{\sfdefault}{bx}{n}

\def\gA{{\mathcal{A}}}

\def\gS{{\mathcal{S}}}

\DeclareMathOperator*{\argmax}{arg\,max}